\documentclass[12pt,letterpaper]{article}
\usepackage[a4paper, total={7in, 10in}]{geometry}

\usepackage{graphicx}
\usepackage{helvet}
\usepackage{authblk}
\usepackage{hyperref}
\usepackage{amsmath} 
\usepackage{amssymb} 
\usepackage{orcidlink} 
\usepackage[super,comma,sort&compress]  
   {natbib}

\usepackage[T1]{fontenc}
\usepackage{multirow}
\usepackage{amsfonts}
\usepackage{amsthm}
\usepackage{xcolor}
\usepackage{booktabs}
\usepackage{csquotes}
\usepackage{tikz}
\usepackage{subcaption}
\usepackage{makecell}
\usepackage{tabularx}
\usepackage{array}
\usepackage{setspace}

\makeatletter
\renewcommand{\maketitle}{\bgroup\setlength{\parindent}{0pt}
\begin{flushleft}
  \textbf{\@title}
  
  \@author
\end{flushleft}\egroup}
\makeatother

\title{What competing constraints reveal about representation learning}
\date{}

\author[1,*,\orcidlink{0000-0003-1669-7140}]{Gwenol\'e Quellec}

\affil[1]{LaTIM UMR 1101, Inserm, Brest, 29200, France}

\affil[*]{Correspondence: gwenole.quellec@inserm.fr}

\begin{document}

\maketitle

\section*{SUMMARY}

Learning latent states from temporal, multimodal, and partially observed data requires decisions about what information representations should retain, discard, and organize. I argue that six complementary properties help clarify these decisions: predictive sufficiency, minimality, temporal coherence, observation compatibility, nuisance invariance, and structural constraints. Drawing on established approaches, I distinguish desired representation properties from the objectives used to encourage them and the diagnostics used to assess them, since neither a training objective nor an evaluation metric guarantees that a property has been achieved. Controlled synthetic experiments examine how training objectives influence predictive performance, nuisance decodability, and latent organization. The results reveal discrepancies between seemingly favorable diagnostics and the information retained in learned representations, as well as latent-coordinate ambiguities under equivalent predictive performance, illustrating how complementary evaluations can improve interpretation and guide design choices. Together, these analyses provide a basis for designing, comparing, and evaluating learned latent states.

\section*{KEYWORDS}

representation learning, latent state modeling, identifiability, multi-objective optimization, self-supervised learning

\section*{INTRODUCTION}

Learning representations from high-dimensional observations is central to machine learning for temporal, multimodal, and partially observed systems. In these settings, a representation may serve as a latent state: a summary of available information intended to support prediction and reasoning about an evolving system, rather than compression alone. State-oriented representations have a long history in predictive processing and are prominent in sequential learning and world models.\cite{rao_predictive_1999,hafner_learning_2019,ha_world_2018} I focus on settings in which this interpretation is useful, drawing on biomedical examples without restricting the argument to a particular application domain.

Existing approaches nevertheless differ in what they encourage a representation to retain or discard. Probabilistic latent-variable models,\cite{kingma_auto-encoding_2014,rezende_stochastic_2014} autoencoders,\cite{hinton_reducing_2006,vincent_extracting_2008} predictive self-supervised methods,\cite{oord_representation_2019,assran_self-supervised_2023} and multimodal architectures\cite{ngiam_multimodal_2011,radford_learning_2021} embody different assumptions about useful latent information. Their downstream performance does not, by itself, reveal which representation properties have been learned. This ambiguity matters in biomedical settings, where acquisition conditions, population differences, and annotation practices can introduce nuisance variation and distribution shifts.\cite{matta_systematic_2024}

I argue that latent-state design can be examined through six complementary properties: predictive sufficiency, minimality, temporal coherence, observation compatibility, nuisance invariance, and structural constraints. These concern, respectively, information retained about relevant futures, unnecessary information discarded, consistency of latent evolution, correspondence with observations, sensitivity to irrelevant variation, and the organization of latent coordinates or geometry. They are neither independent nor exhaustive, and their importance depends on the application. Predictive sufficiency, for example, depends on which future targets and horizons are relevant, rather than being an unqualified property of a representation. Similarly, nuisance invariance must be defined relative to what information is relevant; removing nuisance-associated variation indiscriminately may also remove useful signal.

The properties can interact in different ways. Retaining information needed for prediction may compete with compression, as explored in information-bottleneck approaches.\cite{alemi_deep_2017} Suppressing nuisance variation may conflict with sensitivity to meaningful changes, a concern in self-supervised and contrastive learning.\cite{tschannen_mutual_2020,henaff_data-efficient_2020} Yet these tensions are not universal: the same requirements may also reinforce one another, depending on prediction targets, available information, model class, and training procedure. Multi-objective optimization offers ways to balance specified losses,\cite{sener_multi-task_2018} but choosing weights is different from deciding which representation properties matter and how to assess them.

Viewed in this light, familiar methods emphasize different aspects of representation design rather than occupying mutually exclusive categories. Reconstruction-based approaches encourage observation fidelity without necessarily enforcing future relevance;\cite{alain_what_2014} predictive methods such as contrastive predictive coding emphasize future-related information without guaranteeing predictive sufficiency;\cite{oord_representation_2019} and multimodal learning seeks useful relationships across sources, with varying treatment of nuisance factors and structure.\cite{baltrusaitis_multimodal_2019} Domain-specific models may instead impose stronger structural assumptions, as illustrated by models of disease progression.\cite{fonteijn_event-based_2012,young_uncovering_2018} These emphases describe objectives and inductive biases, not properties that every trained model necessarily satisfies or lacks.

Identifiability makes this distinction especially important. Different latent representations can support equivalent observations or predictions while differing in their coordinates and interpretation. Results on unsupervised disentanglement show why additional inductive biases or supervision are needed to restrict such ambiguities;\cite{locatello_challenging_2019} mutual-information-based objectives alone do not uniquely determine latent structure.\cite{tschannen_mutual_2020} Additional constraints may reduce representational ambiguities, but their ability to establish physical identifiability depends on the data-generating process and modeling assumptions.

Previous work has discussed desirable representation properties and inductive biases,\cite{bengio_representation_2013} reviewed state-learning objectives and evaluation strategies,\cite{lesort_state_2018} translated robotic priors into training objectives,\cite{jonschkowski_learning_2015} and provided guidance on method selection for unsupervised state learning.\cite{overloper_unsupervised_2026} Building on these contributions, this Perspective examines the relationships among three levels: theoretical properties sought in a representation, training surrogates or architectural choices intended to encourage them, and independent diagnostics used to assess selected outcomes. A low loss need not establish the intended property, and a property may arise even when it was not explicitly targeted. The same architecture may support different properties under different prediction targets or data-generating conditions. Likewise, a favorable diagnostic reflects what that measurement can detect, rather than certifying sufficiency, invariance, or interpretability. I examine how discrepancies among these three levels can reveal limitations of learned representations and inform design choices.

To illustrate these relationships, I use a controlled synthetic system. Empirical Pareto fronts characterize compromises among sampled training configurations, while complementary probes reveal information that individual metrics can miss. A target-selection ablation shows how these findings can motivate a design revision evaluated on held-out data. Finally, latent reparameterization and cross-seed alignment distinguish predictive equivalence, coordinate preferences, and empirical reproducibility.

\section*{DESIGN PRINCIPLES FOR LATENT STATE REPRESENTATIONS}

A latent representation can have predictable dynamics while retaining little information about the underlying system, or reconstruct observations accurately while preserving unwanted nuisance variation. These examples illustrate why no single property is sufficient to assess a representation. The six properties should therefore be considered jointly, while distinguishing their theoretical meanings from the training objectives used to encourage them.

Throughout, $T\in\mathbb{N}$ denotes the final time index, $x_t\in\mathcal{X}$ the observation at time $t\in\{1,\ldots,T\}$, and $z_t\in\mathcal{Z}$ the corresponding latent state. The notation $x_{a:b}\in\mathcal{X}^{b-a+1}$ denotes an observation sequence with $1\leq a\leq b\leq T$. For Euclidean representations, $\mathcal{Z}=\mathbb{R}^{d_z}$, where $d_z\geq1$ is the latent dimension. Conditional distributions are denoted by $p(\cdot\mid\cdot)$ and expectations by $\mathbb{E}$. Expressions involving future observations assume $1\leq t<T$.


\subsection*{Predictive sufficiency}

A predictively sufficient latent state retains the information from the observation history needed to characterize the relevant future. For the full sequence of future observations, this property can be expressed as:
\begin{equation}
  p(x_{t+1:T} \mid z_t) = p(x_{t+1:T} \mid x_{1:t}).
\end{equation}

A common training objective encourages this property by predicting future observations from the latent state:
\begin{equation}
  \mathcal{L}_{\text{pred}} = - \mathbb{E}\left[ \log p(x_{t+1:T} \mid z_t) \right].
\end{equation}

Predictive sufficiency is a property of conditional distributions, not of the optimization objective itself. Minimizing a predictive surrogate does not guarantee sufficiency, which depends on the prediction targets, horizons, available information, and model class. In particular, a squared-error objective may accurately predict conditional means without preserving the full distribution of possible futures.


\subsection*{Minimality}

Minimality concerns the information a representation can discard without sacrificing its intended predictive role. Among predictively sufficient representations, a minimal state retains no unnecessary information about the observation history. One information-theoretic formulation expresses this preference through:
\begin{equation}
  \mathcal{L}_{\text{min}} = I(z_t ; x_{1:t}),
\end{equation}

where $I(\cdot;\cdot)$ denotes mutual information. For deterministic encoders with continuous, non-atomic latent outputs, this mutual information is infinite; a finite compression objective requires an appropriate stochastic or quantized formulation. Combined with predictive objectives, this formulation relates compression to the preservation of relevant information. Practical regularizers provide different approximations to this principle. They may discourage memorization or redundant encoding, but minimizing representation complexity does not itself establish information-theoretic minimality or predictive sufficiency.


\subsection*{Temporal coherence}

Temporal coherence concerns the evolution of the latent state. Successive representations should follow dynamics that can be described consistently, rather than fluctuate arbitrarily between observations. A common surrogate uses a learned transition function $f$:
\begin{equation}
  \mathcal{L}_{\text{temp}}
  =
  \mathbb{E}\left[
    \left\|
      f(z_t)-z_{t+1}
    \right\|^2
  \right].
\end{equation}

Here, $f:\mathcal{Z}\to\mathcal{Z}$ predicts the next latent state, and $\|\cdot\|$ denotes the Euclidean norm. This objective favors predictable latent transitions, but transition predictability alone does not establish faithful representation of the underlying dynamics. Nearly constant latent codes, for example, may achieve small transition errors while retaining little useful information about the evolving system.


\subsection*{Observation compatibility}

A latent state should also remain compatible with the observations it represents. This need not require exact reconstruction, particularly when observations are noisy or ambiguous. A probabilistic formulation describes observations through a conditional distribution:
\begin{equation}
  x_t \sim p(x_t \mid z_t).
\end{equation}

A corresponding likelihood-based objective is:
\begin{equation}
  \mathcal{L}_{\text{obs}} = - \mathbb{E}\left[ \log p(x_t \mid z_t) \right].
\end{equation}

Such objectives encourage the latent state to retain information needed to explain observations while allowing probabilistic uncertainty. Reconstruction error provides a useful empirical diagnostic in particular settings, but does not by itself establish general probabilistic observation compatibility.


\subsection*{Invariance to nuisance factors}

Depending on the task, a representation may need to suppress variation associated with nuisance factors, such as acquisition conditions or subject identity, while preserving meaningful changes in the underlying system. Denoting nuisance variables by $c_t$, one strong formulation is marginal invariance:
\begin{equation}
  z_t \perp c_t.
\end{equation}

Here, $\perp$ denotes statistical independence. A corresponding information-theoretic objective minimizes dependence between the representation and the nuisance variables:
\begin{equation}
  \mathcal{L}_{\text{inv}} = I(z_t ; c_t).
\end{equation}

Marginal independence is not universally appropriate. When nuisance variables are associated with task-relevant factors, enforcing it may also suppress useful information. If $s_t$ denotes the relevant physical state, conditional invariance, $z_t \perp c_t \mid s_t$, provides an alternative when the objective is to remove nuisance information beyond that associated with the physical state. Neither formulation is universally preferable: the appropriate choice depends on the data-generating process and the intended use of the representation. Invariance must therefore be considered alongside predictive sufficiency and observation compatibility.


\subsection*{Structural constraints}

Structural constraints concern how information is organized within the latent space, rather than only which information is retained. They can express preferences for smoothness, ordering, manifold structure, factorization, symmetry, equivariance, or relational organization. Other examples include topology- or geometry-preserving representations and domain-specific structural priors. These requirements may be encouraged through regularization terms, denoted here by $\mathcal{L}_{\text{struct}}$, or incorporated directly into model architectures. Their effects depend on the chosen parameterization: constraints that favor particular coordinates or geometric arrangements can distinguish representations with otherwise equivalent predictive behavior. Such preferences can reduce representational ambiguity without necessarily recovering uniquely identifiable physical factors.

\section*{TRADE-OFFS AND DESIGN CHOICES}

The six properties provide complementary, but neither independent nor exhaustive, perspectives on representation design. Their relative importance and interactions depend on the application and modeling assumptions. Some properties can reinforce one another, while others compete under particular training formulations without being incompatible in principle.

Predictive sufficiency and minimality illustrate this distinction. A representation can, in principle, retain the information needed to characterize relevant futures while discarding information unnecessary for that purpose. In practice, stronger compression may impair prediction because of limitations in the representation, predictor, or training objective. Minimality can also support nuisance invariance: Achille and Soatto\cite{achille_emergence_2018} show that minimal sufficient representations are maximally invariant to nuisance variables under the assumptions of their information-theoretic formulation. However, enforcing marginal invariance may remove useful information when nuisance factors are associated with the underlying state.

Observation compatibility introduces another design choice. Reconstruction objectives often favor retaining observational details that are unnecessary for prediction, potentially working against compression or nuisance suppression. Relaxing reconstruction requirements can favor abstraction, but may also discard information needed to interpret the physical system. The appropriate balance therefore depends on what the representation is expected to support. Figure~\ref{fig:tradeoffs} summarizes selected interactions through a qualitative projection onto predictive sufficiency, minimality, and nuisance invariance; temporal coherence, observation compatibility, and structural constraints provide additional dimensions of this design space.

Even when several properties are jointly satisfied, the representation need not be uniquely determined. Identifiability asks whether observations and modeling assumptions determine latent representations up to a specified class of transformations. Different latent coordinates may support equivalent observations and predictions. Structural constraints can restrict these ambiguities, but whether they identify the underlying physical factors depends on the data-generating process, model class, and remaining symmetries.

These considerations guide the comparison of model families in Table~\ref{tab:comparison}. Each family emphasizes particular properties through its objectives, architecture, or modeling assumptions. Such choices do not guarantee that the intended properties are achieved, and properties may emerge without being explicitly targeted. Comparing approaches therefore requires distinguishing what their representations should preserve, how learning encourages those properties, and what independent evaluation shows they actually retain.

\section*{REPRESENTATION CONSTRAINTS ACROSS MODELS}

Representation-learning approaches are commonly distinguished by their architectures or training paradigms. Yet these categories do not always reveal what information their latent states are intended to retain, discard, or organize. I examine representative model families through the properties they seek to encourage and the objectives or architectural choices used to do so. The taxonomy in Figure~\ref{fig:taxonomy} highlights recurring design patterns across methodological traditions, without imposing strict boundaries between approaches.

Individual methods may combine several of these approaches, and similar architectures can produce different representations depending on their training objectives. The comparison therefore concerns the properties that methods seek to encourage, rather than assuming that these properties are necessarily achieved.

\subsection*{State-space models}

State-space models explicitly distinguish the dynamics of an unobserved system from the observations generated by that system. Their central assumption is that latent states evolve according to a Markovian process. For a continuous-state model, one possible formulation is:
\begin{equation}
  z_{t+1} = f(z_t) + \varepsilon_t, \quad x_t \sim p(x_t \mid z_t),
\end{equation}

where $f$ denotes the latent transition function defined above, $\varepsilon_t\in\mathbb{R}^{d_z}$ denotes additive process noise, and $p(\cdot\mid z_t)$ is a conditional probability distribution over the observation space $\mathcal{X}$.
This general modeling tradition includes linear Gaussian systems such as the Kalman and Kalman--Bucy filters,\cite{kalman_new_1960,kalman_new_1961} hidden Markov models,\cite{baum_maximization_1970,rabiner_tutorial_1989} extended and unscented Kalman filters,\cite{wan_unscented_2000,sun_extended_2008} particle-based inference,\cite{kantas_particle_2015} switching state-space models,\cite{ghahramani_variational_2000,fox_bayesian_2011,he_switching_2023} recurrent switching variants,\cite{linderman_bayesian_2017} and hybrid or neural formulations.\cite{feng_review_2023,hashempoor_gated_2024}

Under the Markov assumption, the hidden state determines the distribution of future evolution. However, this state is not directly observed: prediction from observations must account for uncertainty about its value. Under the state-space model, predictions from the observation history are obtained by integrating over the filtering distribution:
\begin{equation}
  p(x_{t+1:T}\mid x_{1:t})
  =
  \int p(x_{t+1:T}\mid z_t)\,
  p(z_t\mid x_{1:t})\,\mathrm{d}z_t.
\end{equation}

Here, $p(z_t\mid x_{1:t})$ represents uncertainty about the hidden state given the observed history. It is distinct from a deterministic learned representation $r_t=\phi_t(x_{1:t})$, where $\phi_t:\mathcal{X}^t\to\mathcal{Z}$ maps observation histories to latent codes. The generative Markov assumption does not guarantee that such a code preserves all information needed to predict future observations.

State-space models also express temporal coherence through their transition dynamics and observation compatibility through $p(x_t\mid z_t)$. These modeling choices help explain their widespread use in speech recognition,\cite{rabiner_tutorial_1989} animal movement ecology,\cite{patterson_state-space_2008} neuroscience,\cite{paninski_new_2010,he_switching_2023} biochemical networks,\cite{sun_extended_2008} and biomechanics.\cite{han_state-space_2015} Their effectiveness nevertheless depends on how well the transition and observation models capture the process being studied; specifying these components does not establish their empirical adequacy. Recent work further illustrates that probabilistic state-space models can learn dynamically relevant representations without an explicit observation decoder.\cite{hromadka_maximum-likelihood_2026}

Other representation properties depend more strongly on additional assumptions. Latent dimensionality, parameterization, and regularization can favor compact representations, but do not guarantee information-theoretic minimality. Identifiability likewise depends on the observation process and restrictions on the model class, including assumptions about dynamics, switching regimes, or prior distributions.\cite{ghahramani_variational_2000,fox_bayesian_2011,linderman_bayesian_2017,feng_review_2023} Structural organization may be introduced through sparse dynamics, mechanistic models, or domain-specific priors, rather than following automatically from the state-space formulation.

Nuisance invariance is similarly application-dependent. Acquisition conditions, population characteristics, or other nuisance factors can be modeled explicitly, marginalized, or accommodated through hierarchical formulations, but classical state-space models do not generally impose independence between latent states and these variables. Their principal strength is therefore the explicit treatment of dynamics, observations, and uncertainty. The extent to which the resulting representations are compact, invariant, interpretable, or identifiable depends on the additional modeling and learning choices.

\subsection*{Latent variable models without temporal structure}

Latent variable models without explicit temporal structure seek compact descriptions of high-dimensional observations. Their foundations include principal component analysis (PCA),\cite{pearson_lines_1901,hotelling_analysis_1933} factor analysis,\cite{spearman_general_1904,lawley_factor_1971} and broader statistical formulations of latent variable models.\cite{bartholomew_latent_2011} Unlike state-space models, they do not explicitly describe transitions between successive latent variables. Their relationship to observations can be expressed schematically as:
\begin{equation}
  x_t \sim p(x_t \mid z_t).
\end{equation}

This family includes linear dimensionality reduction, probabilistic PCA,\cite{tipping_probabilistic_1999} independent component analysis (ICA),\cite{comon_independent_1994,hyvarinen_independent_2000} and nonlinear approaches such as Gaussian process latent variable models.\cite{lawrence_probabilistic_2005} Although their assumptions differ, these methods primarily organize information within observations rather than explicitly modeling its temporal evolution.

Dimensionality reduction encourages compactness, while reconstruction or likelihood-based objectives favor compatibility with observed data.\cite{tipping_probabilistic_1999,bartholomew_latent_2011,lawrence_probabilistic_2005} These choices provide useful low-dimensional summaries, but compactness in the sense of reduced dimensionality is not equivalent to information-theoretic minimality. Similarly, accurately reconstructing observations does not establish that the representation retains the information needed for future prediction.

These models also illustrate how structural assumptions shape latent representations. PCA imposes orthogonal components, factor analysis specifies a loading structure, ICA seeks statistically independent sources, and Gaussian process latent variable models encourage smooth nonlinear mappings.\cite{comon_independent_1994,hyvarinen_independent_2000,lawrence_probabilistic_2005} Such structure can facilitate interpretation, although its relationship to physical factors depends on the data-generating process. Identifiability is likewise conditional on modeling assumptions: nonlinear source separation can remain ambiguous,\cite{hyvarinen_nonlinear_1999} and related difficulties arise in disentangled representation learning.\cite{locatello_challenging_2019,wang_posterior_2021} Recent results also show that geometric restrictions, such as functional orthogonality, can establish identifiability under specific assumptions on the generative model and latent domain.\cite{simon_unsupervised_2026} Statistical independence between latent components, moreover, should not be confused with invariance to external nuisance variables.

Without an explicit transition model, these approaches do not directly encourage temporal coherence. If applied independently to successive observations, they provide no general guarantee that the resulting latent sequence follows consistent dynamics. Predictive sufficiency is also not explicitly enforced: the training objectives do not guarantee that $p(x_{t+1:T}\mid z_t)=p(x_{t+1:T}\mid x_{1:t})$, although this equality may hold when individual observations contain the information required for prediction. Whether such representations support dynamic inference therefore depends on the system and on how they are subsequently used.

Static latent variable models are particularly useful for dimensionality reduction, exploratory analysis, and discovering structure in observations. Their contribution to latent-state modeling lies chiefly in compact representation and observation-oriented learning, rather than explicit dynamics or nuisance control. They also demonstrate why the structural properties encouraged by a learning method must be distinguished from predictive sufficiency, temporal coherence, and invariance in the representations it produces.

\subsection*{Reconstruction-based representation learning}

Reconstruction-based methods learn latent representations by requiring them to retain information from which observations can be recovered. Building on early work on distributed representations and backpropagation,\cite{rumelhart_learning_1986} neural autoencoders introduced nonlinear dimensionality reduction through an encoder-decoder architecture:\cite{hinton_reducing_2006}
\begin{equation}
  z_t = \phi(x_t), \quad \hat{x}_t = \psi(z_t),
\end{equation}

where $\phi:\mathcal{X}\to\mathcal{Z}$ and $\psi:\mathcal{Z}\to\mathcal{X}$ denote the encoder and decoder, respectively.
This schematic formulation encompasses vanilla autoencoders, denoising\cite{vincent_extracting_2008} and contractive variants,\cite{rifai_contractive_2011} variational autoencoders (VAEs),\cite{kingma_auto-encoding_2014,rezende_stochastic_2014} adversarial autoencoders,\cite{makhzani_adversarial_2016} vector-quantized representations,\cite{van_den_oord_neural_2017} hierarchical models such as ladder VAEs,\cite{sonderby_ladder_2016} and hybrids combining reconstruction with predictive or masking objectives.\cite{chen_context_2023} Their training criteria differ, ranging from reconstruction losses to variational objectives that combine observation likelihood and latent regularization.

The principal property encouraged by reconstruction is observation compatibility: the latent code must preserve information needed to reproduce the input, at least to the extent measured by the chosen loss or likelihood model. In deterministic autoencoders, this relationship is expressed as:
\begin{equation}
  x_t \approx \psi(z_t).
\end{equation}

Bottleneck architectures, latent dimensionality, and regularization can additionally encourage compact representations. VAEs, for example, use latent priors and Kullback--Leibler regularization to constrain the learned encoding.\cite{kingma_auto-encoding_2014,rezende_stochastic_2014} These mechanisms influence representation complexity, but reduced dimensionality or latent regularization does not necessarily imply information-theoretic minimality. Likewise, reconstruction fidelity measures agreement with observations under the adopted model rather than establishing general probabilistic observation compatibility.

Additional objectives can shape the organization and robustness of the latent space. Denoising and contractive autoencoders encourage local stability under particular perturbations,\cite{vincent_extracting_2008,rifai_contractive_2011} while variational and adversarial regularization constrain latent distributions.\cite{kingma_auto-encoding_2014,rezende_stochastic_2014,makhzani_adversarial_2016} Approaches such as $\beta$-VAE further encourage factorized representations.\cite{higgins_beta-vae_2017} Yet local robustness or factorization does not necessarily correspond to invariance to task-specific nuisance variables, nor does it establish that latent dimensions recover identifiable physical factors. These ambiguities are central to results on unsupervised disentanglement\cite{locatello_challenging_2019} and to analyses of posterior collapse and latent non-identifiability in variational models.\cite{wang_posterior_2021}

Reconstruction also differs from predictive sufficiency. Optimizing the reconstruction of $x_t$ does not guarantee that $p(x_{t+1:T}\mid z_t)=p(x_{t+1:T}\mid x_{1:t})$, although reconstruction-based representations may satisfy this property under suitable conditions. Standard autoencoders similarly provide no explicit model of temporal evolution; sequential variants require additional mechanisms to learn or impose temporal coherence. Moreover, when acquisition conditions or other nuisance factors contribute to observed variability, accurate reconstruction may favor their retention in the latent code. Even regularizers that encourage useful local geometric structure\cite{alain_what_2014} do not necessarily align representations with task-relevant dynamics.

Reconstruction-based learning thus exposes a central design choice between preserving observational detail and constructing more selective representations. Stronger bottlenecks or regularization may encourage abstraction but reduce reconstruction fidelity or complicate optimization; weaker restrictions may retain information irrelevant to the intended task. Recent representation autoencoders illustrate how pretrained encoders and learned decoders can combine semantic representations with accurate reconstruction.\cite{zheng_diffusion_2026} Across these architectures,\cite{berahmand_autoencoders_2024} predictive relevance, temporal coherence, and nuisance invariance still depend on additional assumptions and objectives.

\subsection*{Predictive representation learning}

Predictive representation learning shifts the emphasis from reconstructing current observations to retaining information useful for anticipating future observations, states, or latent features. Rooted in predictive coding,\cite{rao_predictive_1999} this principle underlies contrastive predictive coding (CPC),\cite{oord_representation_2019} information-bottleneck approaches,\cite{alemi_deep_2017} and joint embedding predictive architectures (JEPA).\cite{assran_self-supervised_2023,bardes_revisiting_2024} In a formulation that predicts future latent features, the representation and predictor can be written as:
\begin{equation}
  z_t=\phi_t(x_{1:t}),\qquad \hat z_{t+k}=g_k(z_t),
\end{equation}

where $\phi_t:\mathcal{X}^t\to\mathcal{Z}$ encodes the observation history, $g_k:\mathcal{Z}\to\mathcal{Z}$ predicts the latent representation at horizon $k\geq1$, and $\hat{z}_{t+k}$ denotes that prediction.
Other formulations predict raw observations or condition future latent transitions on actions, as in world models developed for simulation, planning, and control.\cite{ha_world_2018,hafner_learning_2019,hafner_dream_2020}

The central objective is to preserve information relevant to future prediction, which can be characterized through the mutual information \(I(z_t; x_{t+1:T})\) between the representation and future observations. In practice, predictive methods optimize tractable surrogates involving particular targets, horizons, or prediction tasks. These objectives can encourage predictive sufficiency without guaranteeing that the latent state preserves the full conditional distribution of future observations. Maximizing predictive information also does not, by itself, establish minimality: information unrelated to the chosen prediction task may remain in the representation unless compression or other restrictions are introduced. Information-bottleneck objectives address this distinction by explicitly balancing predictive relevance and compression.\cite{alemi_deep_2017,oord_representation_2019}

Predictive objectives can also encourage nuisance invariance when nuisance-related variation is irrelevant to the selected future targets. This possibility contributes to their effectiveness in self-supervised representation learning.\cite{oord_representation_2019,henaff_data-efficient_2020,assran_self-supervised_2023} However, persistent nuisance factors may themselves be predictive, and some targets may reward their retention. Whether nuisance information is suppressed therefore depends on the data-generating process and the training objective. Predicting latent features rather than raw observations can further favor abstraction, as illustrated by JEPA-style methods,\cite{assran_self-supervised_2023,bardes_revisiting_2024} although the resulting structure is not necessarily aligned with identifiable physical factors.

The treatment of dynamics and observations also varies across predictive architectures. Predicting future representations encourages temporal dependencies, but does not necessarily impose a consistent one-step transition model or coherent latent trajectories. World models may introduce explicit transition dynamics to support planning and long-horizon prediction.\cite{ha_world_2018,hafner_learning_2019,hafner_dream_2020} Recent work extends JEPA to probabilistic prediction, explicitly modeling uncertainty over future latent states.\cite{huang_vjepa_2026} Observation compatibility also depends on the predictive target: methods that predict observed signals impose different requirements from those trained exclusively in embedding space. Neither approach automatically preserves every aspect of the observations relevant to interpretation or mechanistic analysis.

Finally, predictive objectives provide limited control over latent geometry unless additional structural assumptions are introduced. Mutual-information estimation and contrastive optimization can be challenging in high-dimensional settings, and their objectives do not uniquely determine interpretable or disentangled representations.\cite{poole_variational_2019,tschannen_mutual_2020} Predictive learning has nevertheless become a major approach to self-supervised representation learning and latent dynamics modeling.\cite{uelwer_survey_2025} Its central design question is which aspects of the future should be predicted, since that choice influences what the learned representation retains, discards, and makes accessible to downstream models.

\subsection*{Multimodal representation learning}

Multimodal representation learning addresses how information from different sources, such as images, physiological signals, text, audio, or clinical measurements, can be combined within shared or coordinated representations. Beyond concatenating features, these approaches seek to capture relationships between modalities, exploit complementary information, and support cross-modal transfer.\cite{baltrusaitis_multimodal_2019,song_multi-modal_2023,li_review_2024} A generic fusion formulation is:
\begin{equation}
  z_t = \Phi(x_t^{(1)}, \dots, x_t^{(M)}),
\end{equation}

where $M\geq1$ is the number of modalities, $x_t^{(m)}\in\mathcal{X}_m$ denotes the observation from modality $m\in\{1,\ldots,M\}$, and $\Phi:\mathcal{X}_1\times\cdots\times\mathcal{X}_M\to\mathcal{Z}$ combines their information into a latent representation.
The underlying methods differ considerably, from early shared-feature architectures\cite{ngiam_multimodal_2011} and multimodal generative models\cite{srivastava_multimodal_2014,suzuki_joint_2017,wu_multimodal_2018,shi_variational_2019} to deep canonical correlation analysis,\cite{andrew_deep_2013} models of unaligned multimodal sequences,\cite{tsai_multimodal_2019} contrastive vision-language learning such as CLIP and ALIGN,\cite{radford_learning_2021,jia_scaling_2021} and architectures such as ViLBERT and Flamingo.\cite{lu_vilbert_2019,alayrac_flamingo_2022}

These approaches illustrate why modality integration should be distinguished from the properties of the resulting representation. Generative multimodal models can encourage observation compatibility by learning to explain or reconstruct observations from multiple sources.\cite{srivastava_multimodal_2014,suzuki_joint_2017,wu_multimodal_2018} Correlation-based and contrastive methods instead encourage statistical correspondence or alignment between modalities. Such alignment can support retrieval and transfer without requiring that the latent representation explain the full distribution of observations. Thus, combining several modalities does not, by itself, establish probabilistic observation compatibility.

Multimodal learning can nevertheless improve the information available for inference. One modality may resolve ambiguities in another, while shared representations can capture relationships that are difficult to recover from individual sources. These benefits are particularly relevant to vision-language learning and multimodal sequence modeling.\cite{radford_learning_2021,tsai_multimodal_2019,lu_vilbert_2019} Whether the resulting code is predictively sufficient, however, depends on the future targets and information available at inference time. Cross-modal prediction or alignment may encourage useful representations, but neither necessarily preserves all information required to characterize the future. Temporal coherence likewise requires attention to dynamics beyond the integration of observations, even when the inputs are sequential.

Nuisance invariance is another potential benefit, but not an automatic consequence of multimodal alignment. When nuisance variation is specific to one modality, learning shared information may reduce its influence. Conversely, acquisition conditions, population characteristics, or other confounding factors may affect several modalities simultaneously, allowing their information to persist in a shared representation. Contrastive vision-language objectives, for example, can favor semantic correspondence,\cite{radford_learning_2021,jia_scaling_2021} but do not establish independence from task-specific nuisances. Recent multimodal self-supervised work also highlights how strict cross-modal alignment may suppress useful information unique to individual modalities,\cite{cheong_fairssl_2026} underscoring the need to determine what should be shared according to the application.

Integrating multiple sources also raises questions of minimality and latent structure. Complementary modalities may enrich a representation, but fusion can preserve redundant or irrelevant information unless the objective or architecture encourages its removal. Some multimodal generative models explicitly distinguish shared and modality-specific components,\cite{shi_variational_2019} offering a means of organizing latent information. Such decompositions do not necessarily identify the underlying physical factors, and their interpretation remains dependent on modeling assumptions and evaluation.

The main design challenge in multimodal representation learning is therefore not simply how to combine observations, but which relationships across modalities should be preserved and for what purpose. Reconstruction, alignment, prediction, and structured fusion favor different aspects of the latent representation. Distinguishing these objectives from the properties they encourage makes it possible to assess whether additional modalities improve the information relevant to the task, introduce unwanted variability, or change the organization of the learned state.

\subsection*{Domain-specific structured models}

Domain-specific structured models incorporate knowledge of the underlying system into the definition or organization of latent variables. In biomedical applications, these variables may represent disease stages, physiological processes, cognitive states, or treatment responses. Examples include disease progression models, pharmacokinetic--pharmacodynamic (PK/PD) models, and mechanistic or semi-mechanistic approaches.\cite{jack_tracking_2013,chan_drug_2001,dahl_incorporating_2010,zeghlache_latim_2024} Unlike methods that primarily discover latent structure through generic learning objectives, these approaches use domain knowledge to specify what aspects of the system should be represented.

Disease progression modeling illustrates this principle. Event-based models infer an ordering of biomarker abnormalities,\cite{fonteijn_event-based_2012,young_data-driven_2014} while Subtype and Stage Inference (SuStaIn) extends this approach to capture distinct progression patterns and disease heterogeneity.\cite{young_uncovering_2018} Related models describe neurodegeneration through latent factors\cite{zhang_bayesian_2016} or align individuals along estimated disease trajectories.\cite{donohue_estimating_2014,schiratti_learning_2015} Joint models of longitudinal measurements and time-to-event outcomes similarly relate latent processes or classes to clinical trajectories and events.\cite{proust-lima_joint_2014,proust-lima_joint_2016,li_dynamic_2019}

Mechanistic and semi-mechanistic approaches introduce a different form of structure by encoding hypothesized relationships among biological processes, physiological variables, and treatment effects.\cite{clairambault_modelling_2009,mari_mathematical_2020,gaitonde_comprehensive_2016} Hybrid approaches combine such assumptions with data-driven components, including patient-specific predictive models and digital twins.\cite{lorenzo_patient-specific_2024,khoshfekr_rudsari_digital_2025,hussain_neural_2021} Across these methods, structural assumptions determine not only how a model learns, but also which interpretations its latent variables are intended to support.

Structural constraints are therefore central to this family. Disease stages, progression orderings, physiological parameters, and mechanistic relationships provide explicit organization that can facilitate scientific interpretation and hypothesis testing. Some models also encourage temporal coherence through ordered progression, monotonicity assumptions, or explicit dynamical equations.\cite{fonteijn_event-based_2012,young_data-driven_2014,chen_learning_2026} Observation compatibility is commonly addressed through statistical observation models or likelihoods, while prognostic objectives can encourage the retention of information relevant to future clinical outcomes.\cite{proust-lima_joint_2014,young_data-driven_2014} The properties supported by these choices nevertheless depend on the particular model: an ordered disease trajectory, for example, is not equivalent to a general model of individual physiological dynamics.

The principal design choice concerns the balance between domain knowledge and representational flexibility. Strong priors can make latent variables more interpretable and constrain otherwise ambiguous solutions, but incorrect or incomplete assumptions may obscure unexpected trajectories, interactions, or heterogeneity. Moreover, assigning clinical meaning to latent dimensions does not establish their physical identifiability. Low dimensionality does not guarantee information-theoretic minimality, and nuisance invariance generally requires additional attention to acquisition conditions, population differences, and confounding factors. Hybrid models offer ways to combine mechanistic structure with flexible learning, although the resulting representations still require independent assessment of their predictive, observational, and structural properties.

\section*{FROM MODEL COMPARISON TO REPRESENTATION DESIGN}

Across these model families, similar representation goals can be pursued through different learning mechanisms, while similar architectures may emphasize different properties depending on their objectives. Table~\ref{tab:comparison} summarizes these design priorities without treating them as guarantees about the resulting representations.

Several patterns emerge from this comparison. Observation-oriented approaches may preserve details unnecessary for prediction, whereas predictive objectives can favor abstraction without preserving every aspect of the observations. Explicit dynamical models impose temporal assumptions that static representations generally lack, while structural priors organize latent variables according to domain knowledge. Nuisance invariance, meanwhile, depends particularly on the information selected by the training objectives and on the relationship between nuisance variables and task-relevant factors.

These differences across model families turn the comparison into a design question: which properties matter for the task, and which modeling assumptions, architectures, or training objectives can encourage them? Figure~\ref{fig:pipeline} illustrates how the six properties can guide the construction of latent states from observations.

Consider observations $x_{1:T}$, potentially high-dimensional or multimodal, and corresponding latent representations $z_{1:T}$ in a space $\mathcal{Z}$. When modeling an evolving system, latent dynamics may be expressed as:
\begin{equation}
  z_{t+1} = f(z_t) + \varepsilon_t,
\end{equation}

where $f$ denotes the latent transition function defined above, and $\varepsilon_t\in\mathbb{R}^{d_z}$ denotes additive process noise. A probabilistic observation model may additionally relate latent states to measurements:
\begin{equation}
  x_t \sim p(x_t \mid z_t).
\end{equation}

These familiar assumptions express temporal coherence and observation generation through the model structure. Other representation-learning approaches rely instead on discriminative objectives, static encoders, or different probabilistic assumptions. The choice of formulation depends on the properties relevant to the task.

When desired properties can be expressed as restrictions on admissible models, learning can be formulated as a constrained optimization problem:
\begin{equation}
  \min_{\theta} \;\; \mathcal{L}_{\text{task}}(\theta)
  \quad \text{subject to} \quad
  \theta \in \mathcal{C},
\end{equation}

Here, $\theta\in\Theta$ collects the quantities being optimized: model parameters and, where applicable, latent variables, with $\Theta$ denoting the space of candidate solutions. The task objective is $\mathcal{L}_{\text{task}}:\Theta\to\mathbb{R}$, and $\mathcal{C}\subseteq\Theta$ is the set of admissible solutions. If the six properties are defined as explicit constraints, their joint satisfaction can be written as:
\begin{equation}
  \mathcal{C} = \mathcal{C}_{\text{pred}} \cap \mathcal{C}_{\text{min}} \cap \mathcal{C}_{\text{temp}} \cap \mathcal{C}_{\text{obs}} \cap \mathcal{C}_{\text{inv}} \cap \mathcal{C}_{\text{struct}}.
\end{equation}

Each $\mathcal{C}_i\subseteq\Theta$ denotes the set of solutions satisfying the corresponding property under the assumptions of the task. The intersection represents joint satisfaction of the constraints and may be empty if they are incompatible.
Moreover, these properties are not always available as tractable mathematical constraints. Learning procedures therefore commonly employ surrogate objectives or architectural restrictions. One conventional implementation combines training objectives through non-negative weights:
\phantomsection
\label{sec:multi-objective-formulation}
\begin{equation}
  \label{eq:total_loss}
  \mathcal{L}_{\text{total}} =
  \lambda_{\text{pred}} \mathcal{L}_{\text{pred}}
+ \lambda_{\text{min}} \mathcal{L}_{\text{min}}
+ \lambda_{\text{temp}} \mathcal{L}_{\text{temp}}
+ \lambda_{\text{obs}} \mathcal{L}_{\text{obs}}
+ \lambda_{\text{inv}} \mathcal{L}_{\text{inv}}
+ \lambda_{\text{struct}} \mathcal{L}_{\text{struct}}.
\end{equation}

The weights $\lambda_i\in[0,\infty)$, for $i\in\{\text{pred},\text{min},\text{temp}, \text{obs},\text{inv},\text{struct}\}$, control the relative contributions of the selected losses, whose effective influence also depends on their numerical scales. This scalarization is one way to implement multi-objective learning; individual properties may instead be encouraged through probabilistic assumptions, architectural design, or explicit constraints.

Optimizing these surrogates does not guarantee predictive sufficiency, information-theoretic minimality, or identification of physical factors. Conversely, these properties may emerge without being explicitly targeted.

Independent diagnostics therefore provide a complementary level of analysis. Predictive performance, latent-transition consistency, observation fidelity, nuisance decodability, and structural organization can be examined separately from the objectives used during training. Discrepancies between intended properties and measured outcomes may then inform changes to prediction targets, regularization, or architecture. The central design question is not merely how to balance objectives, but how those choices influence the information and structure preserved by learned latent states.

\section*{EXPLORING REPRESENTATION TRADE-OFFS}

To illustrate how intended representation properties, training objectives, and independently measured outcomes can diverge, I use a controlled synthetic benchmark in which the physical state, nuisance variables, and observation process are known.


\subsection*{Controlled latent dynamical environment}

The benchmark simulates a point mass moving within a bounded two-dimensional domain. Its physical state at time $t$ is
\begin{equation}
  s_t=(x_t,y_t,v_{x,t},v_{y,t}),
\end{equation}

where $(x_t,y_t)\in\mathcal{D}\subset\mathbb{R}^2$ denotes position within the bounded spatial domain $\mathcal{D}$, and $(v_{x,t},v_{y,t})\in\mathbb{R}^2$ denotes velocity. Thus, $s_t\in\mathcal{D}\times\mathbb{R}^2$.
The system follows approximately Newtonian dynamics, with small stochastic velocity perturbations and reflections at the domain boundaries. Rather than observing the physical state directly, the model receives a 12-dimensional measurement vector. For each episode $e$, seven nuisance parameters remain fixed throughout the trajectory but vary across episodes: position scales $\alpha_x^{(e)}$ and $\alpha_y^{(e)}$, position offsets $\delta_x^{(e)}$ and $\delta_y^{(e)}$, and sensor biases $b_1^{(e)},b_2^{(e)},b_3^{(e)}$. Each in-distribution (ID) trajectory belongs to one of four uniformly sampled nominal nuisance conditions. These conditions define different distributions over the continuous nuisance parameters and provide a four-class categorical nuisance label. The observation vector is generated as
\begin{equation}
  \begin{bmatrix}
    \alpha_x^{(e)} x_t^{(e)}+\delta_x^{(e)} \\
    \alpha_y^{(e)} y_t^{(e)}+\delta_y^{(e)} \\
    v_{x,t}^{(e)} \\
    v_{y,t}^{(e)} \\
    \sin(\pi x_t^{(e)}) \\
    \cos(\pi y_t^{(e)}) \\
    x_t^{(e)} y_t^{(e)} \\
    (x_t^{(e)})^2-(y_t^{(e)})^2 \\
    b_1^{(e)} \\
    b_2^{(e)} \\
    b_3^{(e)} \\
    1
  \end{bmatrix}
  + \eta_t^{(e)},
  \quad
  \eta_t^{(e)}
  =
  \left(
    \eta_{t,1}^{(e)},\ldots,
    \eta_{t,12}^{(e)}
  \right)^\top,
  \quad
  \eta_{t,j}^{(e)}
  \overset{\mathrm{i.i.d.}}{\sim}
  \mathcal{N}(0,\sigma_{\mathrm{obs}}^2).
\end{equation}

Here, $\eta_t^{(e)}\in\mathbb{R}^{12}$ is the additive observation-noise vector, $j\in\{1,\ldots,12\}$ indexes measurement channels, and $\sigma_{\mathrm{obs}}>0$ is the noise standard deviation. The episode superscript is omitted when unambiguous.

The observation process is deliberately redundant. Velocity is measured directly up to noise, while position is represented through both nuisance-dependent affine measurements and nuisance-independent nonlinear features. Six informative channels---$v_{x,t}$, $v_{y,t}$, $\sin(\pi x_t)$, $\cos(\pi y_t)$, $x_t y_t$, and $x_t^2-y_t^2$---have noiseless signals independent of the episode-level nuisance parameters; their additive measurement noise is also nuisance-independent. Consequently, the model can access physical-state information without relying on nuisance-dependent channels. This design makes it possible to investigate whether training objectives nevertheless encourage the retention of persistent observation-specific variation.

In the noiseless setting, these six channels determine the physical state almost everywhere on the simulated domain. Ambiguities occur only in measure-zero cases, notably at $x_t=0$, where states differing by the sign of $y_t$ cannot be distinguished, and at boundary cases where $\sin(\pi x_t)=0$. Thus, a nuisance-independent representation of the physical state is available in principle, although observation noise prevents this construction from establishing exact statistical sufficiency in the measured data.

Counterfactual consistency is evaluated using paired observations generated from the same physical trajectory under different nuisance conditions. Because the physical states are known, these pairs allow direct measurement of changes in the learned representation when nuisance factors are altered. They are used exclusively for evaluation, never for training.

Table~\ref{tab:toy_dataset} summarizes the dataset. The out-of-distribution (OOD) split preserves the ID latent dynamics, process noise, and observation noise, but changes the nuisance distributions. Position offsets are centered at $\pm1.5$, compared with centers spanning $[-0.8,0.8]$ in the ID splits; position scales and sensor-bias distributions are also shifted and broadened. The OOD evaluation therefore isolates changes in the observation process rather than changes in the underlying dynamics. Unless otherwise stated, results are averaged over five independent random initializations.


\subsection*{Instantiating the constraints}

In the synthetic benchmark, I examine the six representation properties through corresponding differentiable training objectives. Predictive sufficiency is encouraged through multi-step prediction of future observations, while observation compatibility is promoted by reconstructing the current observation. Temporal coherence is encouraged by predicting successive latent states with a learned transition model, and an $\ell_1$ penalty on latent activations encourages smaller code magnitudes as a practical surrogate for minimality. Nuisance invariance is encouraged through adversarial training with a gradient-reversal layer:\cite{ganin_unsupervised_2015} a classifier learns to predict the nuisance condition from the latent representation, while the encoder receives a reversed gradient. Following Ganin and Lempitsky's gradual adaptation principle,\cite{ganin_unsupervised_2015} I use a simple linear warmup of the gradient-reversal coefficient from 0 to 1 over the first 20 epochs. Structural regularization combines a variance-floor penalty with off-diagonal covariance regularization, inspired by VICReg.\cite{bardes_vicreg_2022}

The models use deterministic four-dimensional latent representations. The encoder, decoder, and residual latent-transition model are multilayer perceptrons with two 64-unit hidden layers, ReLU activations, and layer normalization. The encoder maps the 12-dimensional observations to latent states, while the decoder maps latent states back to observation space. When adversarial invariance is enabled, a separate 64--64 ReLU classifier predicts the four nuisance classes. Training and optimization settings are summarized in Table~\ref{tab:hyperparameters}.

Training objectives and evaluation diagnostics are kept separate. Nuisance suppression, for example, is encouraged by the adversarial objective but assessed through counterfactual consistency and independently trained nuisance-decoding probes. This distinction allows the evaluation to reveal nuisance information that may remain accessible even when the training adversary performs poorly. The other representation properties are likewise examined through complementary measurements rather than inferred directly from their training losses.

To explore interactions among objectives, I instantiate Eq.~(\ref{eq:total_loss}) across 200 weight configurations. These comprise 20 reference configurations, including predefined anchors and pilot-derived reference points; 120 configurations sampled locally around the pilot-derived points; and 60 configurations exploring the broader predefined weight ranges. Positive weights are sampled log-uniformly, within local windows for targeted configurations and across the full ranges for exploratory configurations. Fixed quotas introduce zero-valued weights, including settings in which one or two objectives are disabled. Each configuration is evaluated across five independent model seeds. The sampling procedure and experimental settings are detailed in Table~\ref{tab:hyperparameters}.


\subsection*{Constraint interactions in a controlled system}

I assess the learned representations using five complementary post-hoc diagnostics, distinct from the objectives used during training. Together, they characterize predictive performance, physical-state accessibility, latent geometry, counterfactual consistency, and nuisance information. Each measures a particular aspect of the representation rather than directly establishing the corresponding theoretical property.

Multi-step prediction is evaluated through rollout mean squared error (MSE) on the twelve raw observation channels, at multiple horizons, with $h=5$ used for the main Pareto analysis (\textit{Raw-observation prediction MSE, $h=5$, $\downarrow$}). Because several channels depend on persistent episode-level nuisance variables, this metric reflects performance on the chosen observation-space targets rather than nuisance-independent prediction of physical dynamics. To characterize this dependence, I analytically derive the minimum expected raw-observation MSE under conditional nuisance invariance. For each fixed physical state, I calculate the variance in each observation channel caused by the random nuisance parameters: position scales, offsets, and sensor biases. I then average these variances over the twelve channels and the physical states reached at $h=5$. The resulting nuisance-only lower bound is approximately $0.08847$, increasing to $0.08937$ when measurement noise is included, in agreement with an independent Monte Carlo calculation. This bound is specific to raw-observation prediction under conditional nuisance invariance; it is neither the full Bayes prediction error nor a lower bound for predictors that retain episode-specific nuisance information.

To assess physical-state information independently of raw-observation prediction performance, I measure its accessibility through the variance-weighted $R^2$ of a linear probe predicting $s_t$ from $z_t$ (\textit{Linear state $R^2$, $\uparrow$}). As an independent check on the benchmark itself, I also trained a nonlinear probe directly on the six nuisance-independent measurement channels. Despite observation noise, the physical state remained highly accessible ($R^2\approx0.948$), confirming that these channels contain substantial physical-state information without relying on episode-level nuisance variables. This result does not establish exact statistical sufficiency under noisy observations.

Two additional diagnostics examine the organization and stability of the learned representation. Neighborhood preservation is measured by the trustworthiness statistic of Venna and Kaski,\cite{venna_neighborhood_2001} using 15 nearest neighbors to compare local relationships in latent space and physical-state space (\textit{Neighborhood trustworthiness, $\uparrow$}). Counterfactual consistency is evaluated by the relative latent energy between paired observations of the same physical trajectory generated under different nuisance conditions (\textit{Counterfactual relative energy, $\downarrow$}). Lower energy indicates greater consistency under the intervention, although the paired observations also contain independently sampled measurement noise.

Residual nuisance information is assessed using independently trained nonlinear classifiers predicting the four nuisance classes. The primary sweep metric is balanced accuracy from the latent representation alone (\textit{Nuisance $Z$ balanced accuracy, $\downarrow$}). For selected configurations, I additionally evaluate a probe using the joint input $(S,Z)$ (\textit{Nuisance $(S,Z)$ balanced accuracy, $\downarrow$}), alongside a state-only control. Here, $S$ and $Z$ denote the random variables corresponding to the physical state $s_t$ and learned representation $z_t$, respectively. Comparing these probes helps identify nuisance information that is more readily decoded when the physical state is available. Chance-level balanced accuracy is $25\%$. Low nuisance decodability is desirable only when useful information is preserved: a collapsed representation could also yield poor nuisance classification. I therefore interpret these results alongside state accessibility, prediction, neighborhood preservation, and counterfactual consistency. Because nuisance conditions remain fixed within episodes, the 50,000 test frames from 1,000 held-out episodes are not independent. Balanced accuracy measures frame-level nuisance decodability, while variability across model seeds does not capture episode-sampling uncertainty.

I use the five diagnostics to construct pairwise empirical Pareto fronts over the 200 evaluated weight configurations (Figure~\ref{fig:pareto_fronts}). Pareto fronts and configuration selection use validation data; held-out test performance is reported separately below. These fronts characterize the compromises observed within the sampled design rather than universal incompatibilities among representation properties. Some relationships are largely compatible: panel~(D), for example, shows that greater linear state accessibility is generally associated with better neighborhood preservation. Relationships involving raw-observation prediction also depend on the chosen targets, particularly their nuisance-dependent components. The nuisance-related axes quantify marginal nuisance decodability or counterfactual consistency; joint $(S,Z)$ probes provide a complementary examination of nuisance leakage for selected configurations, without establishing exact conditional independence.

I assessed the sensitivity of the empirical Pareto fronts to random initialization by recomputing each front after excluding one of the five training seeds. Across the four metric pairs, mean Jaccard similarity ranged from 0.697 to 0.779, with 80.0--90.0\% of original-front memberships retained across the five leave-one-seed-out analyses (Table~S1). Front membership was therefore partly stable, although individual configurations remained seed-dependent. Representative Pareto-optimal configurations are identified as P1, P2, \ldots, consistently across panels and tables. Their constraint weights and panel memberships are reported in Table~\ref{tab:pareto-weights}.

Table~\ref{tab:main-results} extends the evaluation of representative Pareto configurations and single-constraint anchors with three diagnostics. Current-observation reconstruction MSE (\textit{Obs. recon. MSE, $\downarrow$}) measures observation fidelity, while one-step latent-transition MSE (\textit{1-step latent trans. MSE}) characterizes the predictability of successive latent states. Because latent scales differ across independently trained configurations, these transition errors are not directly comparable between models. A small value may also reflect low latent variability rather than faithful representation of the underlying dynamics. Mean absolute latent activation (\textit{Mean $|z|$}) describes the magnitude of the learned codes, complementing the $\ell_1$ minimality surrogate without measuring information-theoretic minimality. As an uncentered statistic, it cannot, on its own, distinguish a representation collapsed to a nonzero constant from a non-collapsed representation. The table also reports raw-observation prediction MSE and linear physical-state $R^2$ on the OOD test split. No configuration performs best across all reported diagnostics.

The complementary diagnostics reveal differences that individual training objectives can conceal. The temporal-only model has a one-step latent-transition MSE of $7.84\times10^{-6}$ in its own latent coordinates, alongside a linear physical-state $R^2$ of only 0.0916. Although these metrics cannot establish whether the representation has collapsed, their combination illustrates why a small transition error does not establish that the physical state is accessible to a linear decoder. The reconstruction-only model achieves an observation MSE of $0.0126$, while nuisance balanced accuracy from $Z$ reaches $99.0\%$. Accurate reconstruction can therefore coexist with substantial nuisance information. Similarly, mean absolute latent activation measures code magnitude rather than information-theoretic minimality, and reconstruction MSE characterizes observation fidelity in this benchmark rather than general probabilistic observation compatibility.

Nuisance diagnostics reveal a particularly informative discrepancy. For P29, nuisance balanced accuracy rises from $27.2\%$ using $Z$ alone to $52.6\%$ using $(S,Z)$, despite low counterfactual relative energy ($0.0183$). P34 has lower joint nuisance decodability ($34.6\%$), but worse observation-prediction MSE ($0.437$ versus $0.0931$). These comparisons show why marginal nuisance decoding, joint decoding, and counterfactual consistency provide complementary information. None individually establishes conditional nuisance invariance.

The discrepancy observed for P29 motivated a target-selection ablation to assess the contribution of nuisance-dependent prediction targets to nuisance leakage (Table~S2). Restricting prediction to six nuisance-independent channels, while retaining all twelve input channels and adversarial regularization, reduced joint nuisance balanced accuracy from $52.60\pm2.46\%$ to $25.88\pm0.45\%$ (chance: $25\%$). Five-step prediction MSE on these six channels improved both in-distribution ($0.02016$ to $0.01381$) and out-of-distribution ($0.03180$ to $0.02240$), while nonlinear future-state predictability remained high ($R^2$: $0.954$ to $0.941$). Linear physical-state accessibility nevertheless decreased. Without adversarial regularization, nuisance decodability varied substantially across seeds, suggesting that target restriction alone does not reliably suppress nuisance information in this setting.

Synthetic leakage controls further examined the sensitivity of the nuisance probes (Figure~S1). At $\alpha=0.01$, state-modulated nuisance information was more readily decoded from $(S,Z)$ than from $Z$ alone ($37.79\%$ versus $26.82\%$ balanced accuracy). These controls support the interpretation of the decoding experiments without establishing exact conditional independence. The target-selection ablation also relies on knowledge of the synthetic observation process and changes the effective normalization of the predictive loss. Within these limits, the results illustrate a concrete progression from identifying a discrepancy between objectives and diagnostics to revising a design choice and evaluating its consequences on held-out data.


\subsection*{Emergent latent organization}

Complementing the quantitative diagnostics, Figure~\ref{fig:environment_representation} uses known physical trajectories and counterfactual observations to visualize how configuration P29 organizes physical-state information and temporal structure.

The two observation streams differ substantially under the nuisance intervention (Figure~\ref{fig:environment_representation}A), whereas the learned representations exhibit an organization associated with physical position (Figure~\ref{fig:environment_representation}B). Counterfactual observations show substantial overlap in the common PCA projection (Figure~\ref{fig:environment_representation}C), consistent with the low counterfactual relative energy reported for P29. An individual trajectory also follows a smooth, temporally ordered path in this projection (Figure~\ref{fig:environment_representation}D).

These patterns illustrate how predictive, temporal, and invariance-related objectives can jointly shape latent organization without prescribing a particular geometry. Yet the same P29 representation retains nuisance information detectable by the joint $(S,Z)$ probe, despite its counterfactual consistency. The visualization and quantitative diagnostics therefore provide complementary views of the learned state: one reveals its geometric organization, while the other exposes information that may not be apparent in a low-dimensional projection.

\section*{DISCUSSION}

\subsection*{Identifiability and reproducibility}

The coherent latent organization illustrated in Figure~\ref{fig:environment_representation} does not establish that its coordinates are uniquely determined. More generally, different representations may explain the same observations or support equivalent predictions without recovering the same underlying factors. Unsupervised disentanglement generally requires additional inductive biases or supervision to restrict these ambiguities,\cite{locatello_challenging_2019} while reconstruction and mutual-information-based objectives do not uniquely determine latent coordinates.\cite{tschannen_mutual_2020,poole_variational_2019} Structural assumptions can favor particular representations, but whether they establish identifiability depends on the data-generating process, model class, and transformations regarded as equivalent. SplitJEPA provides a recent example of how specific assumptions about predictive dynamics and latent structure can establish identifiability of invariant and variant subspaces, up to defined transformations, without guaranteeing recovery of individual physical factors.\cite{hua_splitjepa_2026} Other recent work distinguishes consistency across independently trained representations from alignment with underlying generative factors,\cite{nelson_statistical_2026} and shows that identifiability metrics can be misleading when their assumptions about the data-generating process or encoder are violated.\cite{joshi_who_2026}

The controlled benchmark illustrates how predictive equivalence can coexist with coordinate-dependent regularization (Figure~\ref{fig:identifiability}). I apply orthogonal reparameterizations to the four-dimensional latent space of a fixed P29 model, transforming its decoder and transition model consistently. As expected from the algebraic invariance under consistent reparameterization, reconstruction, five-step prediction, and latent-transition errors remain unchanged (Figure~6C). This panel provides a numerical consistency check rather than independent empirical evidence. In contrast, the coordinate-dependent minimality-surrogate and structural penalties vary with rotation angle (Figure~6D). Thus, regularization can favor particular coordinate systems even when their predictive behavior is equivalent. These transformations illustrate coordinate ambiguity in a single trained model. I then examine the reproducibility of latent representations across five independently trained P29 models (Supplementary Figure~S2). Transformations fitted on validation data and evaluated on held-out test data reduced the normalized latent mismatch by 77.1\% on average across ten seed pairs using orthogonal alignment. Allowing an additional scaling factor produced a similar reduction of 78.3\%. This substantial but incomplete agreement suggests that independently learned representations share aspects of their organization despite differences in coordinates and optimization outcomes.

These experiments distinguish three questions that are often conflated: whether representations support equivalent predictions, whether regularization favors particular coordinates, and whether independent training runs recover similar latent organizations. None establishes that the learned coordinates correspond uniquely to physical factors. Identifiability must therefore be examined relative to the available observations, modeling assumptions, and admissible transformations, rather than inferred from predictive performance or visually coherent latent geometry.


\subsection*{Interpreting the controlled experiments}

The synthetic benchmark was designed to make representation properties observable under controlled conditions, rather than to reproduce the full difficulty of latent-state inference in real systems. A single observation already contains substantial information about the current physical state, so the experiments do not strongly test inference under partial observability or the integration of long observation histories. In return, access to the physical state, nuisance variables, and counterfactual observations makes it possible to examine aspects of learned representations that are difficult to assess in real-world data.

The empirical Pareto fronts illustrate how different objectives influence representation learning within the sampled configurations. They do not establish that the underlying properties are universally incompatible, particularly since some requirements may be jointly satisfied under suitable assumptions. Their sensitivity to training seeds also highlights the importance of distinguishing persistent patterns from configuration-specific results. More generally, varying the loss weights shows how the chosen training objectives affect learned representations. The resulting trade-offs need not imply conflicts between the theoretical properties those objectives are intended to encourage.

A more informative use of these experiments is to examine discrepancies between intended properties and independently measured outcomes. The P29 results, for example, show that low marginal nuisance decodability and consistent counterfactual representations can coexist with substantial nuisance information accessible jointly with the physical state. Restricting the prediction targets to nuisance-independent channels then provides a concrete example of using such a diagnostic to reconsider a design choice. The resulting improvements in nuisance suppression and selected prediction metrics are informative, although the intervention relies on knowledge of the simulator and also changes the effective normalization of the predictive loss.

These observations place the emphasis on representation design rather than objective weighting alone. Weighted sums are a conventional way to combine training losses, but constrained optimization, adaptive weighting, and Pareto-based approaches provide alternatives when objectives compete. No optimization strategy eliminates the need to decide which properties matter for an application, how they should be encouraged, and how their realization should be evaluated. An accurately optimized surrogate may still favor information that is irrelevant or undesirable for the intended representation.


\subsection*{Implications for design and evaluation}

Representation design therefore starts by specifying what information to preserve, what variation to suppress, and what temporal or structural relationships to express. These requirements guide the choice of prediction targets, architectures, and training objectives. Predictive self-supervised methods, for example, emphasize information useful for selected targets,\cite{oord_representation_2019,assran_self-supervised_2023} but those targets need not capture every aspect of the system relevant to subsequent analysis. Reconstruction, invariance, and structural regularization introduce other priorities whose appropriateness depends on the task.

Evaluation should follow the same reasoning: representations with similar predictive accuracy may differ in nuisance sensitivity, temporal consistency, physical-state accessibility, or structural organization. Complementary diagnostics help reveal these differences, subject to their own limitations. Reconstruction error, latent-transition error, nuisance decodability, and geometric measurements each probe selected aspects of a representation rather than certifying the corresponding theoretical properties.

This distinction is especially relevant to clinical applications. Two models with comparable diagnostic or prognostic accuracy may differ in their sensitivity to acquisition conditions, population shifts, or changes in the underlying physiological process. Their latent variables may also differ in stability and interpretability. Such differences matter when representations are used to study disease trajectories, integrate longitudinal measurements, or support clinical decisions. Evaluating these characteristics alongside task performance can therefore provide a more informative basis for model selection.

The challenge becomes greater as representation learning scales. Large self-supervised and multimodal models demonstrate the effectiveness of learning from heterogeneous, high-dimensional data,\cite{radford_learning_2021,alayrac_flamingo_2022,openai_gpt-4_2024} but their complexity makes it difficult to characterize the information retained in internal representations. Explicitly evaluating temporal behavior, nuisance sensitivity, or structural organization may require specialized probes and substantial computational resources. Scalable diagnostics and careful target selection are consequently important complements to advances in architecture and optimization.

Beyond the computational demands of evaluation, real-world systems may also require representation properties beyond the six examined here. Biological processes, for example, operate across interacting temporal and spatial scales, from rapid physiological fluctuations to disease progression over years. Hierarchical representations may help relate these processes, while causal assumptions, mechanistic models, and geometry-aware constraints can express additional knowledge about the system. Dynamic causal modeling\cite{friston_dynamic_2003} and patient-specific mechanistic modeling\cite{lorenzo_patient-specific_2024} illustrate how domain assumptions can guide the organization of latent variables. Their usefulness depends on the validity of those assumptions, and additional structure does not automatically establish causal or physical identifiability.

A further direction is to develop evaluation protocols that compare representation properties more systematically across learning paradigms. Benchmarks could examine prediction at specified horizons, robustness under controlled distribution shifts, consistency of latent dynamics, nuisance leakage, and stability of latent organization across training runs. Where ground truth is unavailable, these evaluations would necessarily rely on partial diagnostics rather than direct verification of theoretical properties. Reusable implementations of training objectives, probes, and evaluation procedures could nevertheless facilitate more reproducible comparisons and help identify which design choices matter across applications.

\section*{CONCLUSION}

I have examined latent-state learning through six complementary and interacting representation properties, rather than through architectures or training paradigms alone. Existing approaches differ in the information they seek to preserve, discard, and organize. Their objectives and inductive biases can encourage desirable properties, but these properties must be distinguished from the training surrogates used to promote them and the diagnostics used to evaluate them.

The controlled experiments illustrate why this distinction matters. Empirical trade-offs depend on the chosen objectives and prediction targets, while complementary diagnostics can reveal information that individual metrics overlook. The target-selection ablation shows how such findings can inform representation design, and the identifiability analyses distinguish predictive equivalence, coordinate preferences, and reproducibility across training runs.

The central question is therefore not which architecture or training objective is universally preferable, but which representation properties matter for a given task, how learning choices encourage them, and how their realization can be assessed independently. Connecting these three levels provides a more precise basis for designing, interpreting, and comparing learned latent states.


\newpage


\section*{RESOURCE AVAILABILITY}


\subsection*{Lead contact}


Requests for further information and resources should be directed to and will be fulfilled by the lead contact, Gwenol\'e Quellec (gwenole.quellec@inserm.fr).

\subsection*{Materials availability}


This study did not generate new materials.

\subsection*{Data and code availability}


A companion repository containing the point-mass simulator and data-generation code, the reference implementation and training scripts, reproducible experiments and documentation is available at: \href{https://github.com/gwenole-quellec/clsm}{https://github.com/gwenole-quellec/clsm}.
The corresponding release (v0.3.0, commit \texttt{c38b57e}) is archived on Zenodo (\url{https://doi.org/10.5281/zenodo.24142180}).

\section*{ACKNOWLEDGMENTS}


This work was supported by the French National Research Agency (ANR) through the projects TEDIA (ANR-24-CE45-1133) and EvoluPark (ANR-25-PESN-0005), as part of the France 2030 program.

\section*{AUTHOR CONTRIBUTIONS}


Conceptualization, methodology, software, formal analysis, investigation, visualization, writing--original draft, writing--review \& editing, funding acquisition, G.Q.

\section*{DECLARATION OF INTERESTS}


The author declares no competing interests.

\section*{DECLARATION OF GENERATIVE AI AND AI-ASSISTED TECHNOLOGIES}


During the preparation of this work, the author used ChatGPT (OpenAI) to assist with developing the code for the synthetic benchmark and improving the clarity and concision of the manuscript. After using this tool or service, the author reviewed and edited the content as needed and takes full responsibility for the content of the publication.

\section*{SUPPLEMENTAL INFORMATION INDEX}




\begin{description}
  \item{Table S1.} Leave-one-seed-out stability of empirical pairwise Pareto fronts.
  \item{Table S2.} Effect of prediction targets and adversarial invariance.
  \item{Figure S1.} Sensitivity of nuisance probes to synthetic nuisance leakage.
  \item{Figure S2.} Cross-seed reproducibility of learned latent representations.
\end{description}

\newpage

\bibliography{references}

@article{kalman_new_1960,
	title = {A new approach to linear filtering and prediction problems},
	volume = {82},
	issn = {0021-9223},
	doi = {10.1115/1.3662552},
	number = {1},
	urldate = {2026-04-19},
	journal = {J Basic Eng},
	author = {Kalman, R. E.},
	month = mar,
	year = {1960},
	pages = {35--45},
}

@article{kalman_new_1961,
	title = {New results in linear filtering and prediction theory},
	volume = {83},
	issn = {0021-9223},
	doi = {10.1115/1.3658902},
	number = {1},
	urldate = {2026-04-19},
	journal = {J Basic Eng},
	author = {Kalman, R. E. and Bucy, R. S.},
	month = mar,
	year = {1961},
	pages = {95--108},
}

@article{rabiner_tutorial_1989,
	title = {A tutorial on hidden {Markov} models and selected applications in speech recognition},
	volume = {77},
	issn = {1558-2256},
	doi = {10.1109/5.18626},
	number = {2},
	urldate = {2026-04-19},
	journal = {Proc IEEE},
	author = {Rabiner, L.R.},
	month = feb,
	year = {1989},
	pages = {257--286},
}

@article{baum_maximization_1970,
	title = {A maximization technique occurring in the statistical analysis of probabilistic functions of {Markov} chains},
	volume = {41},
	issn = {0003-4851},
	doi = {10.1214/aoms/1177697196},
	number = {1},
	urldate = {2026-04-19},
	journal = {Ann Math Stat},
	publisher = {Institute of Mathematical Statistics},
	author = {Baum, Leonard E. and Petrie, Ted and Soules, George and Weiss, Norman},
	month = feb,
	year = {1970},
	pages = {164--171},
}

@article{patterson_state-space_2008,
	title = {State-space models of individual animal movement},
	volume = {23},
	issn = {0169-5347},
	doi = {10.1016/j.tree.2007.10.009},
	number = {2},
	journal = {Trends Ecol Evol},
	author = {Patterson, Toby A. and Thomas, Len and Wilcox, Chris and Ovaskainen, Otso and Matthiopoulos, Jason},
	month = feb,
	year = {2008},
	pages = {87--94},
}

@article{paninski_new_2010,
	title = {A new look at state-space models for neural data},
	volume = {29},
	issn = {1573-6873},
	doi = {10.1007/s10827-009-0179-x},
	number = {1-2},
	journal = {J Comput Neurosci},
	author = {Paninski, Liam and Ahmadian, Yashar and Ferreira, Daniel Gil and Koyama, Shinsuke and Rahnama Rad, Kamiar and Vidne, Michael and Vogelstein, Joshua and Wu, Wei},
	month = aug,
	year = {2010},
	pages = {107--126},
}

@inproceedings{wan_unscented_2000,
	location = {Lake Louise, AB, Canada},
	title = {The unscented {Kalman} filter for nonlinear estimation},
	doi = {10.1109/ASSPCC.2000.882463},
	urldate = {2026-04-20},
	booktitle = {Proc {IEEE} {AS}-{SPCC}},
	publisher = {IEEE},
    editor = {Haykin, Simon and Principe, Jos{\'e} C.},
	author = {Wan, E. A. and van der Merwe, R.},
	month = oct,
	year = {2000},
	pages = {153--158},
}

@article{kantas_particle_2015,
	title = {On particle methods for parameter estimation in state-space models},
	volume = {30},
	issn = {0883-4237, 2168-8745},
	doi = {10.1214/14-STS511},
	number = {3},
	urldate = {2026-04-19},
	journal = {Stat Sci},
	publisher = {Institute of Mathematical Statistics},
	author = {Kantas, Nikolas and Doucet, Arnaud and Singh, Sumeetpal S. and Maciejowski, Jan and Chopin, Nicolas},
	month = aug,
	year = {2015},
	pages = {328--351},
}

@article{sun_extended_2008,
	title = {Extended {Kalman} filter for estimation of parameters in nonlinear state-space models of biochemical networks},
	volume = {3},
	issn = {1932-6203},
	doi = {10.1371/journal.pone.0003758},
	number = {11},
	urldate = {2026-04-19},
	journal = {PLOS ONE},
	publisher = {Public Library of Science},
	author = {Sun, Xiaodian and Jin, Li and Xiong, Momiao},
	month = nov,
	year = {2008},
	pages = {e3758},
}

@article{ghahramani_variational_2000,
	title = {Variational learning for switching state-space models},
	volume = {12},
	issn = {0899-7667},
	doi = {10.1162/089976600300015619},
	number = {4},
	journal = {Neural Comput},
	author = {Ghahramani, Z. and Hinton, G. E.},
	month = may,
	year = {2000},
	pages = {831--864},
}

@article{he_switching_2023,
	title = {Switching state-space modeling of neural signal dynamics},
	volume = {19},
	issn = {1553-7358},
	doi = {10.1371/journal.pcbi.1011395},
	number = {8},
	urldate = {2026-04-19},
	journal = {PLoS Comput Biol},
	publisher = {Public Library of Science},
	author = {He, Mingjian and Das, Proloy and Hotan, Gladia and Purdon, Patrick L.},
	month = aug,
	year = {2023},
	pages = {e1011395},
}

@article{feng_review_2023,
	title = {A review: {State} estimation based on hybrid models of {Kalman} filter and neural network},
	volume = {11},
	shorttitle = {A review},
	doi = {10.1080/21642583.2023.2173682},
	number = {1},
	urldate = {2026-04-19},
	journal = {Syst Sci Control Eng},
	publisher = {Taylor \& Francis},
	author = {Feng, Shuo and Li, Xuegui and Zhang, Shuai and Jian, Zhen and Duan, Hanxu and Wang, Zepeng},
	month = dec,
	year = {2023},
	pages = {2173682},
}

@inproceedings{hashempoor_gated_2024,
	title = {Gated inference network: {Inference} and learning state-space models},
	volume = {37},
    location = {Vancouver, BC, Canada},
	shorttitle = {Gated inference network},
	doi = {10.52202/079017-1232},
	urldate = {2026-04-19},
	booktitle = {Adv {Neural} {Inf} {Process} {Syst}},
	author = {Hashempoor, Hamidreza and Choi, Wan},
    publisher = {Curran Associates, Inc.},
    editor = {Globerson, A. and Mackey, L. and Belgrave, D. and Fan, A. and Paquet, U. and Tomczak, J. and Zhang, C.},
	month = dec,
	year = {2024},
	pages = {39036--39073},
}

@article{han_state-space_2015,
	title = {A state-space {EMG} model for the estimation of continuous joint movements},
	volume = {62},
	issn = {1557-9948},
	doi = {10.1109/TIE.2014.2387337},
	number = {7},
	urldate = {2026-04-19},
	journal = {IEEE Trans Ind Electron},
	author = {Han, Jianda and Ding, Qichuan and Xiong, Anbin and Zhao, Xingang},
	month = jul,
	year = {2015},
	pages = {4267--4275},
}

@article{fox_bayesian_2011,
	title = {Bayesian nonparametric inference of switching dynamic linear models},
	volume = {59},
	issn = {1053-587X},
	doi = {10.1109/TSP.2010.2102756},
	number = {4},
	urldate = {2026-04-19},
	journal = {IEEE Trans Sig Proc},
	author = {Fox, E. and Sudderth, E. B. and Jordan, M. I. and Willsky, A. S.},
	month = apr,
	year = {2011},
	pages = {1569--1585},
}

@inproceedings{linderman_bayesian_2017,
	title = {Bayesian learning and inference in recurrent switching linear dynamical systems},
	issn = {2640-3498},
    location = {Fort Lauderdale, FL, USA},
	urldate = {2026-04-19},
	booktitle = {Proc {AISTATS}},
	publisher = {PMLR},
    editor = {Singh, Aarti and Zhu, Jerry},
    volume = {54},
	author = {Linderman, Scott and Johnson, Matthew and Miller, Andrew and Adams, Ryan and Blei, David and Paninski, Liam},
	month = apr,
	year = {2017},
	pages = {914--922},
    url = {https://proceedings.mlr.press/v54/linderman17a.html},
}

@article{pearson_lines_1901,
	title = {On lines and planes of closest fit to systems of points in space},
	volume = {2},
	issn = {1941-5982},
	doi = {10.1080/14786440109462720},
	number = {11},
	urldate = {2026-04-19},
	journal = {Lond Edinb Dubl Phil Mag},
	publisher = {Taylor \& Francis},
	author = {Pearson, Karl},
	month = nov,
	year = {1901},
	pages = {559--572},
}

@article{hotelling_analysis_1933,
	title = {Analysis of a complex of statistical variables into principal components},
	volume = {24},
	issn = {1939-2176},
	doi = {10.1037/h0071325},
	number = {6},
	journal = {J Educ Psychol},
	publisher = {Warwick \& York},
	author = {Hotelling, H.},
	year = {1933},
	pages = {417--441},
}

@article{spearman_general_1904,
	title = {General intelligence, objectively determined and measured},
	volume = {15},
	issn = {1939-8298},
	doi = {10.2307/1412107},
	number = {2},
	journal = {Am J Psychol},
	publisher = {Univ of Illinois Press},
	author = {Spearman, C.},
	month = apr,
	year = {1904},
	pages = {201--293},
}

@book{lawley_factor_1971,
	title = {Factor analysis as a statistical method},
	isbn = {978-0-408-70152-5},
	urldate = {2026-04-19},
	location = {London, UK},
    publisher = {Butterworths},
	author = {Lawley, D. N. and Maxwell, A. E.},
	collaborator = {{Internet Archive}},
	year = {1971},
}

@book{bartholomew_latent_2011,
	address = {Chichester, UK},
	title = {Latent variable models and factor analysis: {A} unified approach},
	isbn = {978-0-470-97192-5},
	shorttitle = {Latent variable models and factor analysis},
	publisher = {Wiley},
	author = {Bartholomew, David J. and Knott, Martin and Moustaki, Irini},
	month = jun,
	year = {2011},
}

@article{tipping_probabilistic_1999,
	title = {Probabilistic principal component analysis},
	volume = {61},
	copyright = {1999 Royal Statistical Society},
	issn = {1467-9868},
	doi = {10.1111/1467-9868.00196},
	number = {3},
	urldate = {2026-04-19},
	journal = {J R Stat Soc, B},
	author = {Tipping, Michael E. and Bishop, Christopher M.},
	year = {1999},
	pages = {611--622},
}

@article{comon_independent_1994,
	series = {Higher {Order} {Statistics}},
	title = {Independent component analysis, a new concept?},
	volume = {36},
	issn = {0165-1684},
	doi = {10.1016/0165-1684(94)90029-9},
	number = {3},
	urldate = {2026-04-19},
	journal = {Signal Process},
	author = {Comon, Pierre},
	month = apr,
	year = {1994},
	pages = {287--314},
}

@article{hyvarinen_independent_2000,
	title = {Independent component analysis: {Algorithms} and applications},
	volume = {13},
	issn = {0893-6080},
	shorttitle = {Independent component analysis},
	doi = {10.1016/S0893-6080(00)00026-5},
	number = {4},
	urldate = {2026-04-19},
	journal = {Neural Netw},
	author = {Hyv{\"a}rinen, A. and Oja, E.},
	month = jun,
	year = {2000},
	pages = {411--430},
}

@article{lawrence_probabilistic_2005,
	title = {Probabilistic non-linear principal component analysis with {Gaussian} process latent variable models},
	volume = {6},
	issn = {1532-4435},
	doi = {10.5555/1046920.1194904},
	urldate = {2026-04-19},
	journal = {J Mach Learn Res},
	author = {Lawrence, Neil},
	month = dec,
	year = {2005},
	pages = {1783--1816},
}

@inproceedings{locatello_challenging_2019,
	location = {Long Beach, CA, USA},
	title = {Challenging common assumptions in the unsupervised learning of disentangled representations},
    author = {Locatello, Francesco and Bauer, Stefan and Lucic, Mario and R{\"a}tsch, Gunnar and Gelly, Sylvain and Sch{\"o}lkopf, Bernhard and Bachem, Olivier},
    booktitle = {Proc ICML},
    pages = {4114--4124},
    year = {2019},
    editor = {Chaudhuri, Kamalika and Salakhutdinov, Ruslan},
    volume = {97},
    month = jun,
    publisher = {PMLR},
    url = {https://proceedings.mlr.press/v97/locatello19a.html}
}

@article{hyvarinen_nonlinear_1999,
	title = {Nonlinear independent component analysis: {Existence} and uniqueness results},
	volume = {12},
	issn = {0893-6080},
	shorttitle = {Nonlinear independent component analysis},
	doi = {10.1016/S0893-6080(98)00140-3},
	number = {3},
	urldate = {2026-04-19},
	journal = {Neural Netw},
	author = {Hyv{\"a}rinen, Aapo and Pajunen, Petteri},
	month = apr,
	year = {1999},
	pages = {429--439},
}

@inproceedings{wang_posterior_2021,
	title = {Posterior collapse and latent variable non-identifiability},
	volume = {34},
	url = {https://papers.neurips.cc/paper_files/paper/2021/hash/2b6921f2c64dee16ba21ebf17f3c2c92-Abstract.html},
	urldate = {2026-04-19},
	booktitle = {Adv {Neural} {Inf} {Process} {Syst}},
	publisher = {Curran Associates, Inc.},
    editor = {Ranzato, Marc'Aurelio and Beygelzimer, Alina and Dauphin, Yann and Liang, Percy S. and Wortman Vaughan, Jennifer},
	author = {Wang, Yixin and Blei, David and Cunningham, John P.},
	year = {2021},
	pages = {5443--5455},
}

@inproceedings{kingma_auto-encoding_2014,
	location = {Banff, AB, Canada},
	title = {Auto-encoding variational {Bayes}},
	urldate = {2026-04-19},
	booktitle = {Proc {ICLR}},
	author = {Kingma, D. P. and Welling, M.},
    editor = {Bengio, Yoshua and LeCun, Yann},
    url = {https://mlanthology.org/iclr/2014/kingma2014iclr-auto/},
	month = apr,
	year = {2014},
}

@inproceedings{rezende_stochastic_2014,
	location = {Beijing, China},
	title = {Stochastic backpropagation and approximate inference in deep generative models},
	doi = {10.5555/3044805.3045035},
	urldate = {2026-04-19},
	booktitle = {Proc {ICML}},
	publisher = {PMLR},
    editor = {Xing, Eric and Jebara, Tony},
	author = {Rezende, Danilo Jimenez and Mohamed, Shakir and Wierstra, Daan},
	month = jun,
	year = {2014},
    volume = {32},
	pages = {1278--1286},
}

@inproceedings{higgins_beta-vae_2017,
	location = {Toulon, France},
	title = {beta-{VAE}: {Learning} basic visual concepts with a constrained variational framework},
	shorttitle = {beta-{VAE}},
	urldate = {2026-04-19},
	booktitle = {Proc {ICLR}},
	url = {https://openreview.net/forum?id=Sy2fzU9gl},
    editor = {Bengio, Yoshua and LeCun, Yann},
	author = {Higgins, Irina and Matthey, Lo{\"i}c and Pal, Arka and Burgess, Christopher P. and Glorot, Xavier and Botvinick, Matthew M. and Mohamed, Shakir and Lerchner, Alexander},
	month = apr,
	year = {2017},
}

@article{rumelhart_learning_1986,
	title = {Learning representations by back-propagating errors},
	volume = {323},
	copyright = {1986 Springer Nature Limited},
	issn = {1476-4687},
	doi = {10.1038/323533a0},
	number = {6088},
	urldate = {2026-04-20},
	journal = {Nature},
	publisher = {Nature Publishing Group},
	author = {Rumelhart, David E. and Hinton, Geoffrey E. and Williams, Ronald J.},
	month = oct,
	year = {1986},
	pages = {533--536},
}

@article{hinton_reducing_2006,
	title = {Reducing the dimensionality of data with neural networks},
	volume = {313},
	doi = {10.1126/science.1127647},
	number = {5786},
	urldate = {2026-04-20},
	journal = {Science},
	publisher = {American Association for the Advancement of Science},
	author = {Hinton, G. E. and Salakhutdinov, R. R.},
	month = jul,
	year = {2006},
	pages = {504--507},
}

@inproceedings{vincent_extracting_2008,
	location = {Helsinki, Finland},
	title = {Extracting and composing robust features with denoising autoencoders},
	isbn = {978-1-60558-205-4},
	doi = {10.1145/1390156.1390294},
	urldate = {2026-04-19},
	booktitle = {Proc {ICML}},
    publisher = {PMLR},
    editor = {Cohen, William and McCallum, Andrew and Roweis, Sam T.},
	author = {Vincent, Pascal and Larochelle, Hugo and Bengio, Yoshua and Manzagol, Pierre-Antoine},
	month = jul,
	year = {2008},
	pages = {1096--1103},
}

@inproceedings{rifai_contractive_2011,
	location = {Bellevue, WA, USA},
	title = {Contractive auto-encoders: {Explicit} invariance during feature extraction},
	isbn = {978-1-4503-0619-5},
	shorttitle = {Contractive auto-encoders},
	doi = {10.5555/3104482.3104587},
	urldate = {2026-04-19},
	booktitle = {Proc {ICML}},
	publisher = {PMLR},
    editor = {Getoor, Lise and Scheffer, Tobias},
	author = {Rifai, Salah and Vincent, Pascal and Muller, Xavier and Glorot, Xavier and Bengio, Yoshua},
	month = jun,
	year = {2011},
	pages = {833--840},
}

@inproceedings{makhzani_adversarial_2016,
	location = {San Juan, Puerto Rico},
	title = {Adversarial autoencoders},
	urldate = {2026-04-20},
	booktitle = {ICLR Workshop Track},
	url = {https://openreview.net/forum?id=2xwp4Zwr3TpKBZvXtWoj},
    editor = {Bengio, Yoshua and LeCun, Yann},
	author = {Makhzani, Alireza and Shlens, Jonathon and Jaitly, Navdeep and Goodfellow, Ian},
	month = may,
	year = {2016},
}

@inproceedings{van_den_oord_neural_2017,
	location = {Long Beach, CA, USA},
	title = {Neural discrete representation learning},
	volume = {30},
	url = {https://papers.neurips.cc/paper_files/paper/2017/hash/7a98af17e63a0ac09ce2e96d03992fbc-Abstract.html},
	urldate = {2026-04-20},
	booktitle = {Adv {Neural} {Inf} {Process} {Syst}},
	publisher = {Curran Associates, Inc.},
    editor = {Guyon, Isabelle and {von Luxburg}, Ulrike and Bengio, Samy and Wallach, Hanna M. and Fergus, Rob and Vishwanathan, S. V. N. and Garnett, Roman},
	author = {van den Oord, A{\"a}ron and Vinyals, Oriol and Kavukcuoglu, Koray},
	month = dec,
	year = {2017},
	pages = {6309--6318},
}

@inproceedings{sonderby_ladder_2016,
	location = {Red Hook, NY, USA},
	title = {Ladder variational autoencoders},
	url = {https://papers.neurips.cc/paper_files/paper/2016/hash/6ae07dcb33ec3b7c814df797cbda0f87-Abstract.html},
	urldate = {2026-04-19},
	booktitle = {Adv {Neural} {Inf} {Process} {Syst}},
	publisher = {Curran Associates, Inc.},
    editor = {Lee, Daniel D and Sugiyama, Masashi and {von Luxburg}, Ulrike and Guyon, Isabelle and Garnett, Roman},
	author = {S{\o}nderby, Casper Kaae and Raiko, Tapani and Maal{\o}e, Lars and Sønderby, S{\o}ren Kaae and Winther, Ole},
	month = dec,
	year = {2016},
	pages = {3745--3753},
}

@article{alain_what_2014,
	title = {What regularized auto-encoders learn from the data-generating distribution},
	volume = {15},
	issn = {1533-7928},
	doi = {10.5555/2627435.2750359},
	number = {110},
	urldate = {2026-04-20},
	journal = {J Mach Learn Res},
	author = {Alain, Guillaume and Bengio, Yoshua},
	month = jan,
	year = {2014},
	pages = {3743--3773},
}

@article{berahmand_autoencoders_2024,
	title = {Autoencoders and their applications in machine learning: {A} survey},
	volume = {57},
	issn = {1573-7462},
	shorttitle = {Autoencoders and their applications in machine learning},
	doi = {10.1007/s10462-023-10662-6},
	number = {2},
	urldate = {2026-04-20},
	journal = {Artif Intell Rev},
	author = {Berahmand, Kamal and Daneshfar, Fatemeh and Salehi, Elaheh Sadat and Li, Yuefeng and Xu, Yue},
	month = feb,
	year = {2024},
	pages = {28},
}

@article{chen_context_2023,
	title = {Context autoencoder for self-supervised representation learning},
	volume = {132},
	issn = {0920-5691},
	doi = {10.1007/s11263-023-01852-4},
	number = {1},
	urldate = {2026-04-20},
	journal = {Int J Comput Vision},
	author = {Chen, Xiaokang and Ding, Mingyu and Wang, Xiaodi and Xin, Ying and Mo, Shentong and Wang, Yunhao and Han, Shumin and Luo, Ping and Zeng, Gang and Wang, Jingdong},
	month = jan,
	year = {2024},
	pages = {208--223},
}

@inproceedings{alemi_deep_2017,
	location = {Toulon, France},
	title = {Deep variational information bottleneck},
	booktitle = {Proc {ICLR}},
    editor = {Bengio, Yoshua and LeCun, Yann},
	author = {Alemi, Alex and Fischer, Ian and Dillon, Josh and Murphy, Kevin},
	month = apr,
	year = {2017},
	url = {https://mlanthology.org/iclr/2017/alemi2017iclr-deep/},
}

@article{rao_predictive_1999,
	title = {Predictive coding in the visual cortex: {A} functional interpretation of some extra-classical receptive-field effects},
	volume = {2},
	copyright = {1999 Nature America Inc.},
	issn = {1546-1726},
	shorttitle = {Predictive coding in the visual cortex},
	doi = {10.1038/4580},
	number = {1},
	urldate = {2026-04-20},
	journal = {Nat Neurosci},
	publisher = {Nature Publishing Group},
	author = {Rao, Rajesh P. N. and Ballard, Dana H.},
	month = jan,
	year = {1999},
	pages = {79--87},
}

@article{oord_representation_2019,
	title = {Representation learning with contrastive predictive coding},
	author = {van den Oord, A{\"a}ron and Li, Yazhe and Vinyals, Oriol},
	doi = {10.48550/arXiv.1807.03748},
    journal = {Preprint at arXiv},
	  year = {2018},
}

@inproceedings{poole_variational_2019,
	location = {Long Beach, CA, USA},
	title = {On variational bounds of mutual information},
	issn = {2640-3498},
	urldate = {2026-04-20},
	booktitle = {Proc {ICML}},
	publisher = {PMLR},
    editor = {Chaudhuri, Kamalika and Salakhutdinov, Ruslan},
	author = {Poole, Ben and Ozair, Sherjil and van den Oord, A{\"a}ron and Alemi, Alex and Tucker, George},
	month = may,
	year = {2019},
	pages = {5171--5180},
    volume = {97},
    url = {https://proceedings.mlr.press/v97/poole19a.html},
}

@inproceedings{assran_self-supervised_2023,
	location = {Vancouver, BC, Canada},
	title = {Self-supervised learning from images with a joint-embedding predictive architecture},
	doi = {10.1109/CVPR52729.2023.01499},
	urldate = {2026-04-20},
	booktitle = {Proc {CVPR}},
	publisher = {IEEE},
    editor = {Brown, Michael S. and Li, Fei-Fei and Mori, Greg and Sato, Yoichi},
	author = {Assran, Mahmoud and Duval, Quentin and Misra, Ishan and Bojanowski, Piotr and Vincent, Pascal and Rabbat, Michael and LeCun, Yann and Ballas, Nicolas},
	month = jun,
	year = {2023},
}

@inproceedings{ha_world_2018,
    address = {Montreal, QC, Canada},
	title = {Recurrent world models facilitate policy evolution},
	author = {Ha, David and Schmidhuber, J{\"u}rgen},
    booktitle = {Adv {Neural} {Inf} {Process} {Syst}},
    editor = {Bengio, Samy and Wallach, Hanna M. and Larochelle, Hugo and Grauman, Kristen and Cesa-Bianchi, Nicol{\`o} and Garnett, Roman},
    volume = {31},
    pages = {2451--2463},
    publisher = {Curran Associates, Inc.},
    year = {2018},
    url = {https://papers.neurips.cc/paper_files/paper/2018/hash/2de5d16682c3c35007e4e92982f1a2ba-Abstract.html}
}

@inproceedings{hafner_learning_2019,
	location = {Long Beach, CA, USA},
	title = {Learning latent dynamics for planning from pixels},
	issn = {2640-3498},
	urldate = {2026-04-20},
	booktitle = {Proc {ICML}},
	publisher = {PMLR},
    editor = {Chaudhuri, Kamalika and Salakhutdinov, Ruslan},
	author = {Hafner, Danijar and Lillicrap, Timothy and Fischer, Ian and Villegas, Ruben and Ha, David and Lee, Honglak and Davidson, James},
	month = may,
	year = {2019},
	pages = {2555--2565},
    url = {https://proceedings.mlr.press/v97/hafner19a.html},
}

@inproceedings{hafner_dream_2020,
	title = {Dream to control: {Learning} behaviors by latent imagination},
	shorttitle = {Dream to control},
	urldate = {2026-04-20},
	booktitle = {Proc {ICLR}},
	url = {https://openreview.net/forum?id=S1lOTC4tDS},
	author = {Hafner, Danijar and Lillicrap, Timothy and Ba, Jimmy and Norouzi, Mohammad},
	month = mar,
	year = {2020},
}

@article{bardes_revisiting_2024,
	title = {Revisiting feature prediction for learning visual representations from video},
	urldate = {2026-04-20},
	journal = {Trans Mach Learn Res},
	author = {Bardes, Adrien and Garrido, Quentin and Ponce, Jean and Chen, Xinlei and Rabbat, Michael and LeCun, Yann and Assran, Mahmoud and Ballas, Nicolas},
	month = aug,
	year = {2024},
    issn = {2835-8856},
	url = {https://openreview.net/forum?id=QaCCuDfBk2},
}

@inproceedings{tschannen_mutual_2020,
	title = {On mutual information maximization for representation learning},
	urldate = {2026-04-20},
	booktitle = {Proc {ICLR}},
	author = {Tschannen, Michael and Djolonga, Josip and Rubenstein, Paul K. and Gelly, Sylvain and Lucic, Mario},
	month = may,
	year = {2020},
    url = {https://openreview.net/forum?id=rkxoh24FPH},
}

@inproceedings{henaff_data-efficient_2020,
	title = {Data-efficient image recognition with contrastive predictive coding},
	issn = {2640-3498},
	doi = {10.5555/3524938.3525329},
	urldate = {2026-04-20},
	booktitle = {Proc {ICML}},
	publisher = {PMLR},
    editor = {Daum{\'e}, III, Hal and Singh, Aarti},
	author = {H{\'e}naff, Olivier},
	month = nov,
	year = {2020},
	pages = {4182--4192},
}

@article{uelwer_survey_2025,
	title = {A survey on self-supervised methods for visual representation learning},
	volume = {114},
	issn = {1573-0565},
	doi = {10.1007/s10994-024-06708-7},
	number = {4},
	urldate = {2026-04-20},
	journal = {Mach Learn},
	author = {Uelwer, Tobias and Robine, Jan and Wagner, Stefan Sylvius and H{\"o}ftmann, Marc and Upschulte, Eric and Konietzny, Sebastian and Behrendt, Maike and Harmeling, Stefan},
	month = mar,
	year = {2025},
	pages = {111},
}

@inproceedings{radford_learning_2021,
	title = {Learning transferable visual models from natural language supervision},
	urldate = {2024-02-29},
	booktitle = {Proc {ICML}},
	publisher = {PMLR},
    editor = {Meila, Marina and Zhang, Tong},
	author = {Radford, Alec and Kim, Jong Wook and Hallacy, Chris and Ramesh, Aditya and Goh, Gabriel and Agarwal, Sandhini and Sastry, Girish and Askell, Amanda and Mishkin, Pamela and Clark, Jack and others},
	month = jul,
	year = {2021},
    volume = {139},
	pages = {8748--8763},
    url = {https://proceedings.mlr.press/v139/radford21a.html},
}

@inproceedings{ngiam_multimodal_2011,
	location = {Bellevue, WA, USA},
	title = {Multimodal deep learning},
	booktitle = {Proc {ICML}},
	publisher = {Omnipress},
    editor = {Getoor, Lise and Scheffer, Tobias},
	author = {Ngiam, Jiquan and Khosla, Aditya and Kim, Mingyu and Nam, Juhan and Lee, Honglak and Ng, Andrew Y.},
	month = jul,
	year = {2011},
	pages = {689--696},
    url = {https://icml.cc/Conferences/2011/papers.php.html},
}

@article{srivastava_multimodal_2014,
	title = {Multimodal learning with deep {Boltzmann} machines},
	volume = {15},
	number = {84},
	journal = {J Mach Learn Res},
	author = {Srivastava, Nitish and Salakhutdinov, Ruslan},
	year = {2014},
	pages = {2949--2980},
    url = {https://jmlr.org/papers/v15/srivastava14b.html},
}

@inproceedings{suzuki_joint_2017,
	location = {Toulon, France},
	title = {Joint multimodal learning with deep generative models},
	urldate = {2026-04-20},
	booktitle = {ICLR Workshop Track},
    editor = {Bengio, Yoshua and LeCun, Yann},
	author = {Suzuki, Masahiro and Nakayama, Kotaro and Matsuo, Yutaka},
	month = apr,
	year = {2017},
	url = {https://openreview.net/forum?id=oYvplybLTu},
}

@inproceedings{wu_multimodal_2018,
	location = {Montreal, QC, Canada},
	title = {Multimodal generative models for scalable weakly-supervised learning},
	volume = {31},
	url = {https://papers.neurips.cc/paper_files/paper/2018/hash/1102a326d5f7c9e04fc3c89d0ede88c9-Abstract.html},
	urldate = {2026-04-20},
	booktitle = {Adv {Neural} {Inf} {Process} {Syst}},
	publisher = {Curran Associates, Inc.},
    editor = {Bengio, Samy and Wallach, Hanna M. and Larochelle, Hugo and Grauman, Kristen and Cesa-Bianchi, Nicol{\`o} and Garnett, Roman},
	author = {Wu, Mike and Goodman, Noah},
	month = dec,
	year = {2018},
	pages = {5575--5585},
}

@inproceedings{shi_variational_2019,
	location = {Vancouver, BC, Canada},
	title = {Variational mixture-of-experts autoencoders for multi-modal deep generative models},
	volume = {32},
	url = {https://papers.neurips.cc/paper_files/paper/2019/hash/0ae775a8cb3b499ad1fca944e6f5c836-Abstract.html},
	urldate = {2026-04-20},
	booktitle = {Adv {Neural} {Inf} {Process} {Syst}},
	publisher = {Curran Associates, Inc.},
    editor = {Wallach, Hanna M. and Larochelle, Hugo and Beygelzimer, Alina and {d'Alch{\'e}-Buc}, Florence and Fox, Emily B. and Garnett, Roman},
	author = {Shi, Yuge and Siddharth, N. and Paige, Brooks and Torr, Philip H. S.},
	month = dec,
	year = {2019},
	pages = {15718--15729},
}

@inproceedings{andrew_deep_2013,
	location = {Atlanta, GA, USA},
	title = {Deep canonical correlation analysis},
	url = {https://proceedings.mlr.press/v28/andrew13.html},
	urldate = {2026-04-20},
	booktitle = {Proc {ICML}},
	publisher = {PMLR},
    editor = {Dasgupta, Sanjoy and McAllester, David},
	author = {Andrew, Galen and Arora, Raman and Bilmes, Jeff and Livescu, Karen},
	month = jun,
	year = {2013},
    volume = {28},
    pages = {1247--1255},
}

@inproceedings{jia_scaling_2021,
	title = {Scaling up visual and vision-language representation learning with noisy text supervision},
	volume = {139},
	issn = {2640-3498},
	urldate = {2026-04-20},
	booktitle = {Proc {ICML}},
	publisher = {PMLR},
    editor = {Meila, Marina and Zhang, Tong},
	author = {Jia, Chao and Yang, Yinfei and Xia, Ye and Chen, Yi-Ting and Parekh, Zarana and Pham, Hieu and Le, Quoc and Sung, Yun-Hsuan and Li, Zhen and Duerig, Tom},
	month = jul,
	year = {2021},
	pages = {4904--4916},
    url = {https://proceedings.mlr.press/v139/jia21b.html},
}

@inproceedings{tsai_multimodal_2019,
	location = {Florence, Italy},
	title = {Multimodal transformer for unaligned multimodal language sequences},
	volume = {57},
	doi = {10.18653/v1/P19-1656},
	urldate = {2026-04-20},
	booktitle = {Proc {ACL}},
	publisher = {ACL},
	editor = {Korhonen, Anna and Traum, David and M{\`a}rquez, Llu{\'i}s},
	author = {Tsai, Yao-Hung Hubert and Bai, Shaojie and Liang, Paul Pu and Kolter, J. Zico and Morency, Louis-Philippe and Salakhutdinov, Ruslan},
	month = jul,
	year = {2019},
	pages = {6558--6569},
}

@inproceedings{lu_vilbert_2019,
	title = {{ViLBERT}: {Pretraining} task-agnostic visiolinguistic representations for vision-and-language tasks},
	volume = {32},
	url = {https://papers.neurips.cc/paper_files/paper/2019/hash/c74d97b01eae257e44aa9d5bade97baf-Abstract.html},
    address = {Vancouver, BC, Canada},
	urldate = {2026-04-20},
	booktitle = {Adv {Neural} {Inf} {Process} {Syst}},
	publisher = {Curran Associates, Inc.},
    editor = {Wallach, Hanna M. and Larochelle, Hugo and Beygelzimer, Alina and {d'Alch{\'e}-Buc}, Florence and Fox, Emily B. and Garnett, Roman},
	author = {Lu, Jiasen and Batra, Dhruv and Parikh, Devi and Lee, Stefan},
	month = dec,
	year = {2019},
	pages = {13--23},
}

@inproceedings{alayrac_flamingo_2022,
	title = {{Flamingo}: {A} visual language model for few-shot learning},
	volume = {35},
	shorttitle = {Flamingo},
	doi = {10.52202/068431-1723},
	urldate = {2026-04-20},
	booktitle = {Adv {Neural} {Inf} {Process} {Syst}},
	publisher = {Curran Associates, Inc.},
    editor = {Koyejo, Sanmi and Mohamed, Shakir and Agarwal, Alekh and Belgrave, Danielle and Cho, Kyunghyun and Oh, Alice},
	author = {Alayrac, Jean-Baptiste and Donahue, Jeff and Luc, Pauline and Miech, Antoine and Barr, Iain and Hasson, Yana and Lenc, Karel and Mensch, Arthur and Millican, Katherine and Reynolds, Malcolm and others},
    address = {New Orleans, LA, USA},
	month = dec,
	year = {2022},
	pages = {23716--23736},
}

@article{baltrusaitis_multimodal_2019,
	title = {Multimodal machine learning: {A} survey and taxonomy},
	volume = {41},
	doi = {10.1109/TPAMI.2018.2798607},
	number = {2},
	urldate = {2026-04-20},
	journal = {IEEE Trans Pattern Anal Mach Intell},
	publisher = {IEEE Computer Society},
	author = {Baltrusaitis, Tadas and Ahuja, Chaitanya and Morency, Louis-Philippe},
	month = feb,
	year = {2019},
	pages = {423--443},
}

@article{song_multi-modal_2023,
	title = {Multi-modal machine learning in engineering design: {A} review and future directions},
	volume = {24},
	issn = {1530-9827},
	shorttitle = {Multi-{Modal} {Machine} {Learning} in {Engineering} {Design}},
	doi = {10.1115/1.4063954},
	number = {1},
	urldate = {2026-04-20},
	journal = {J Comput Inf Sci Eng},
	author = {Song, Binyang and Zhou, Rui and Ahmed, Faez},
	month = jan,
	year = {2024},
	pages = {010801},
}

@article{fonteijn_event-based_2012,
	title = {An event-based model for disease progression and its application in familial {Alzheimer}'s disease and {Huntington}'s disease},
	volume = {60},
	issn = {1095-9572},
	doi = {10.1016/j.neuroimage.2012.01.062},
	number = {3},
	journal = {Neuroimage},
	author = {Fonteijn, Hubert M. and Modat, Marc and Clarkson, Matthew J. and Barnes, Josephine and Lehmann, Manja and Hobbs, Nicola Z. and Scahill, Rachael I. and Tabrizi, Sarah J. and Ourselin, Sebastien and Fox, Nick C. and others},
	month = apr,
	year = {2012},
	pages = {1880--1889},
}

@article{young_data-driven_2014,
	title = {A data-driven model of biomarker changes in sporadic {Alzheimer}'s disease},
	volume = {137},
	issn = {1460-2156},
	doi = {10.1093/brain/awu176},
	number = {Pt 9},
	journal = {Brain},
	author = {Young, Alexandra L. and Oxtoby, Neil P. and Daga, Pankaj and Cash, David M. and Fox, Nick C. and Ourselin, Sebastien and Schott, Jonathan M. and Alexander, Daniel C. and {Alzheimer's Disease Neuroimaging Initiative}},
	month = sep,
	year = {2014},
	pages = {2564--2577},
}

@article{young_uncovering_2018,
	title = {Uncovering the heterogeneity and temporal complexity of neurodegenerative diseases with {Subtype} and {Stage} {Inference}},
	volume = {9},
	copyright = {2018 The Author(s)},
	issn = {2041-1723},
	doi = {10.1038/s41467-018-05892-0},
	number = {1},
	urldate = {2026-04-20},
	journal = {Nat Commun},
	publisher = {Nature Publishing Group},
	author = {Young, Alexandra L. and Marinescu, Razvan V. and Oxtoby, Neil P. and Bocchetta, Martina and Yong, Keir and Firth, Nicholas C. and Cash, David M. and Thomas, David L. and Dick, Katrina M. and Cardoso, Jorge and others},
	month = oct,
	year = {2018},
	pages = {4273},
}

@article{proust-lima_joint_2014,
	title = {Joint latent class models for longitudinal and time-to-event data: {A} review},
	volume = {23},
	issn = {1477-0334},
	shorttitle = {Joint latent class models for longitudinal and time-to-event data},
	doi = {10.1177/0962280212445839},
	number = {1},
	journal = {Stat Methods Med Res},
	author = {Proust-Lima, C{\'e}cile and S{\'e}ne, Mb{\'e}ry and Taylor, Jeremy M. G. and Jacqmin-Gadda, H{\'e}l{\`e}ne},
	month = feb,
	year = {2014},
	pages = {74--90},
}

@article{friston_dynamic_2003,
	title = {Dynamic causal modelling},
	volume = {19},
	issn = {1053-8119},
	doi = {10.1016/s1053-8119(03)00202-7},
	number = {4},
	journal = {Neuroimage},
	author = {Friston, K. J. and Harrison, L. and Penny, W.},
	month = aug,
	year = {2003},
	pages = {1273--1302},
}

@article{clairambault_modelling_2009,
	title = {Modelling physiological and pharmacological control on cell proliferation to optimise cancer treatments},
	volume = {4},
	issn = {0973-5348},
	doi = {10.1051/mmnp/20094302},
	number = {3},
	urldate = {2026-04-20},
	journal = {Math Model Nat Phenom},
	publisher = {EDP Sciences},
	author = {Clairambault, J.},
	year = {2009},
	pages = {12--67},
}

@inproceedings{schiratti_learning_2015,
	  location = {Montreal, QC, Canada},
	title = {Learning spatiotemporal trajectories from manifold-valued longitudinal data},
	volume = {28},
	url = {https://proceedings.neurips.cc/paper/2015/hash/186a157b2992e7daed3677ce8e9fe40f-Abstract.html},
	urldate = {2026-04-20},
	booktitle = {Adv {Neural} {Inf} {Process} {Syst}},
	publisher = {Curran Associates, Inc.},
    editor = {Cortes, Corinna and Lawrence, Neil D. and Lee, Daniel D. and Sugiyama, Masashi and Garnett, Roman},
	author = {Schiratti, Jean-Baptiste and Allassonni{\`e}re, St{\'e}phanie and Colliot, Olivier and Durrleman, Stanley},
	month = dec,
	year = {2015},
	pages = {2404--2412},
}

@article{proust-lima_joint_2016,
	title = {Joint modeling of repeated multivariate cognitive measures and competing risks of dementia and death: {A} latent process and latent class approach},
	volume = {35},
	copyright = {Copyright © 2015 John Wiley \& Sons, Ltd.},
	issn = {1097-0258},
	shorttitle = {Joint modeling of repeated multivariate cognitive measures and competing risks of dementia and death},
	doi = {10.1002/sim.6731},
	number = {3},
	urldate = {2026-04-20},
	journal = {Stat Med},
	author = {Proust-Lima, C{\'e}cile and Dartigues, Jean-François and Jacqmin-Gadda, H{\'e}l{\`e}ne},
	year = {2016},
	pages = {382--398},
}

@article{li_dynamic_2019,
	title = {Dynamic predictions in {Bayesian} functional joint models for longitudinal and time-to-event data: {An} application to {Alzheimer}’s disease},
	volume = {28},
	issn = {0962-2802},
	shorttitle = {Dynamic {Predictions} in {Bayesian} {Functional} {Joint} {Models} for {Longitudinal} and {Time}-to-{Event} {Data}},
	doi = {10.1177/0962280217722177},
	number = {2},
	urldate = {2026-04-20},
	journal = {Stat Methods Med Res},
	author = {Li, Kan and Luo, Sheng},
	month = feb,
	year = {2019},
	pages = {327--342},
}

@article{donohue_estimating_2014,
	title = {Estimating long-term multivariate progression from short-term data},
	volume = {10},
	issn = {1552-5260},
	doi = {10.1016/j.jalz.2013.10.003},
	number = {0},
	urldate = {2026-04-20},
	journal = {Alzheimers Dement},
	author = {Donohue, Michael C. and Jacqmin-Gadda, H{\'e}l{\`e}ne and Le Goff, M{\'e}lanie and Thomas, Ronald G. and Raman, Rema and Gamst, Anthony C. and Beckett, Laurel A. and Jack, Clifford R. and Weiner, Michael W. and Dartigues, Jean-Fran{\c{c}}ois and others},
	month = oct,
	year = {2014},
	pages = {S400--S410},
}

@article{jack_tracking_2013,
	title = {Tracking pathophysiological processes in {Alzheimer}'s disease: {An} updated hypothetical model of dynamic biomarkers},
	volume = {12},
	issn = {1474-4422, 1474-4465},
	shorttitle = {Tracking pathophysiological processes in {Alzheimer}'s disease},
	doi = {10.1016/S1474-4422(12)70291-0},
	number = {2},
	urldate = {2026-04-20},
	journal = {Lancet Neurol},
	publisher = {Elsevier},
	author = {Jack, Clifford R. and Knopman, David S. and Jagust, William J. and Petersen, Ronald C. and Weiner, Michael W. and Aisen, Paul S. and Shaw, Leslie M. and Vemuri, Prashanthi and Wiste, Heather J. and Weigand, Stephen D. and others},
	month = feb,
	year = {2013},
	pages = {207--216},
}

@article{zhang_bayesian_2016,
	title = {Bayesian model reveals latent atrophy factors with dissociable cognitive trajectories in {Alzheimer}'s disease},
	volume = {113},
	issn = {1091-6490},
	doi = {10.1073/pnas.1611073113},
	number = {42},
	journal = {Proc Natl Acad Sci USA},
	author = {Zhang, Xiuming and Mormino, Elizabeth C. and Sun, Nanbo and Sperling, Reisa A. and Sabuncu, Mert R. and Yeo, B. T. Thomas and {Alzheimer’s Disease Neuroimaging Initiative}},
	month = oct,
	year = {2016},
	pages = {E6535--E6544},
}

@article{chan_drug_2001,
	title = {Drug treatment effects on disease progression},
	volume = {41},
	issn = {0362-1642},
	doi = {10.1146/annurev.pharmtox.41.1.625},
	journal = {Annu Rev Pharmacol Toxicol},
	author = {Chan, P. L. and Holford, N. H.},
	year = {2001},
	pages = {625--659},
}

@article{dahl_incorporating_2010,
	title = {Incorporating physiological and biochemical mechanisms into pharmacokinetic-pharmacodynamic models: {A} conceptual framework},
	volume = {106},
	issn = {1742-7843},
	shorttitle = {Incorporating physiological and biochemical mechanisms into pharmacokinetic-pharmacodynamic models},
	doi = {10.1111/j.1742-7843.2009.00456.x},
	number = {1},
	journal = {Basic Clin Pharmacol Toxicol},
	author = {Dahl, Svein G. and Aarons, Leon and Gundert-Remy, Ursula and Karlsson, Mats O. and Schneider, Yves-Jacques and Steimer, Jean-Louis and Troc{\'o}niz, I{\~n}aki F.},
	month = jan,
	year = {2010},
	pages = {2--12},
}

@article{gaitonde_comprehensive_2016,
	title = {A comprehensive review of novel drug-disease models in diabetes drug development},
	volume = {55},
	issn = {1179-1926},
	doi = {10.1007/s40262-015-0359-y},
	number = {7},
	journal = {Clin Pharmacokinet},
	author = {Gaitonde, Puneet and Garhyan, Parag and Link, Catharina and Chien, Jenny Y. and Trame, Mirjam N. and Schmidt, Stephan},
	month = jul,
	year = {2016},
	pages = {769--788},
}

@article{mari_mathematical_2020,
	title = {Mathematical modeling for the physiological and clinical investigation of glucose homeostasis and diabetes},
	volume = {11},
	issn = {1664-042X},
	doi = {10.3389/fphys.2020.575789},
	journal = {Front Physiol},
	author = {Mari, Andrea and Tura, Andrea and Grespan, Eleonora and Bizzotto, Roberto},
	year = {2020},
	pages = {575789},
}

@article{lorenzo_patient-specific_2024,
	title = {Patient-specific, mechanistic models of tumor growth incorporating artificial intelligence and big data},
	volume = {26},
	issn = {1545-4274},
	doi = {10.1146/annurev-bioeng-081623-025834},
	number = {1},
	journal = {Annu Rev Biomed Eng},
	author = {Lorenzo, Guillermo and Ahmed, Syed Rakin and Hormuth, David A. and Vaughn, Brenna and Kalpathy-Cramer, Jayashree and Solorio, Luis and Yankeelov, Thomas E. and Gomez, Hector},
	month = jul,
	year = {2024},
	pages = {529--560},
}

@inproceedings{hussain_neural_2021,
	title = {Neural pharmacodynamic state space modeling},
	volume = {139},
	issn = {2640-3498},
	urldate = {2026-04-20},
	booktitle = {Proc {ICML}},
	publisher = {PMLR},
    editor = {Meila, Marina and Zhang, Tong},
	author = {Hussain, Zeshan M. and Krishnan, Rahul G. and Sontag, David},
	month = jul,
	year = {2021},
	pages = {4500--4510},
    url = {https://proceedings.mlr.press/v139/hussain21a.html},
}

@article{khoshfekr_rudsari_digital_2025,
	title = {Digital twins in healthcare: {A} comprehensive review and future directions},
	volume = {7},
	issn = {2673-253X},
	shorttitle = {Digital twins in healthcare},
	doi = {10.3389/fdgth.2025.1633539},
	urldate = {2026-04-20},
	journal = {Front Digit Health},
	author = {Khoshfekr Rudsari, Hamid and Tseng, Becky and Zhu, Hongxu and Song, Lulu and Gu, Chunhui and Roy, Abhishikta and Irajizad, Ehsan and Butner, Joseph and Long, James and Do, Kim-Anh},
	month = nov,
	year = {2025},
	pages = {1633539},
}

@article{openai_gpt-4_2024,
  title = {{GPT}-4 Technical Report},
  author = {{OpenAI}},
  year = {2023},
  journal = {Preprint at arXiv},
  doi = {10.48550/arXiv.2303.08774},
}

@article{matta_systematic_2024,
	title = {A systematic review of generalization research in medical image classification},
	volume = {183},
	issn = {0010-4825},
	doi = {10.1016/j.compbiomed.2024.109256},
	urldate = {2024-11-04},
	journal = {Comput Biol Med},
	author = {Matta, Sarah and Lamard, Mathieu and Zhang, Philippe and Le Guilcher, Alexandre and Borderie, Laurent and Cochener, B{\'e}atrice and Quellec, Gwenol{\'e}},
	month = dec,
	year = {2024},
	pages = {109256},
}

@article{li_review_2024,
	title = {A review of deep learning-based information fusion techniques for multimodal medical image classification},
	volume = {177},
	issn = {0010-4825},
	doi = {10.1016/j.compbiomed.2024.108635},
	urldate = {2025-06-18},
	journal = {Comput Biol Med},
	author = {Li, Yihao and El Habib Daho, Mostafa and Conze, Pierre-Henri and Zeghlache, Rachid and Le Boit{\'e}, Hugo and Tadayoni, Ramin and Cochener, B{\'e}atrice and Lamard, Mathieu and Quellec, Gwenol{\'e}},
	month = jul,
	year = {2024},
	pages = {108635},
}

@inproceedings{zeghlache_latim_2024,
	location = {Marrakech, Morocco},
	title = {{LaTiM}: {Longitudinal} representation learning in continuous-time models to predict disease progression},
	shorttitle = {{LaTiM}},
	doi = {10.1007/978-3-031-72086-4_38},
	urldate = {2024-10-07},
	booktitle = {Proc {MICCAI}},
    series = {LNCS},
    volume = {15005},
	publisher = {Springer},
	editor = {Linguraru, Marius George and Dou, Qi and Feragen, Aasa and Giannarou, Stamatia and Glocker, Ben and Lekadir, Karim and Schnabel, Julia A.},
	author = {Zeghlache, Rachid and Conze, Pierre-Henri and El Habib Daho, Mostafa and Li, Yihao and Le Boit{\'e}, Hugo and Tadayoni, Ramin and Massin, Pascal and Cochener, B{\'e}atrice and Rezaei, Alireza and Brahim, Ikram and others},
	month = oct,
	year = {2024},
	pages = {404--414},
}

@article{achille_emergence_2018,
	title = {Emergence of invariance and disentanglement in deep representations},
	volume = {19},
	issn = {1533-7928},
	url = {https://jmlr.org/papers/v19/17-646.html},
	number = {50},
	urldate = {2026-10-03},
	journal = {Journal of Machine Learning Research},
	author = {Achille, Alessandro and Soatto, Stefano},
	year = {2018},
	pages = {1--34},
}

@inproceedings{bardes_vicreg_2022,
	title = {{VICReg}: {Variance}-invariance-covariance regularization for self-supervised learning},
    author = {Bardes, Adrien and Ponce, Jean and LeCun, Yann},
    editor = {Liu, Yan and Finn, Chelsea and Choi, Yejin and Deisenroth, Marc},
    booktitle = {Proc. ICLR},
    year = {2022},
    url = {https://openreview.net/forum?id=iWpcWZ8phD},
}

@article{bengio_representation_2013,
	title = {Representation learning: {A} review and new perspectives},
	volume = {35},
	issn = {1939-3539},
	shorttitle = {Representation learning},
	doi = {10.1109/TPAMI.2013.50},
	number = {8},
	urldate = {2026-10-03},
	journal = {IEEE Trans Pattern Anal Mach Intell},
	author = {Bengio, Yoshua and Courville, Aaron and Vincent, Pascal},
	month = aug,
	year = {2013},
	pages = {1798--1828},
}

@inproceedings{ganin_unsupervised_2015,
	address = {Lille, France},
	title = {Unsupervised domain adaptation by backpropagation},
	volume = {37},
	doi = {10.5555/3045118.3045244},
	urldate = {2026-10-05},
	booktitle = {Proc {ICML}},
	publisher = {JMLR.org},
	author = {Ganin, Yaroslav and Lempitsky, Victor},
    editor = {Bach, Francis and Blei, David},
	month = jul,
	year = {2015},
	pages = {1180--1189},
}

@article{jonschkowski_learning_2015,
	title = {Learning state representations with robotic priors},
	volume = {39},
	issn = {1573-7527},
	doi = {10.1007/s10514-015-9459-7},
	number = {3},
	urldate = {2026-10-03},
	journal = {Auton Robot},
	author = {Jonschkowski, Rico and Brock, Oliver},
	month = oct,
	year = {2015},
	pages = {407--428},
}

@article{lesort_state_2018,
	title = {State representation learning for control: {An} overview},
	volume = {108},
	issn = {0893-6080},
	shorttitle = {State representation learning for control},
	doi = {10.1016/j.neunet.2018.07.006},
	urldate = {2026-10-03},
	journal = {Neural Networks},
	author = {Lesort, Timoth{\'e}e and D{\'i}az-Rodr{\'i}guez, Natalia and Goudou, Jean-Fran{\c{c}}ois and Filliat, David},
	month = dec,
	year = {2018},
	pages = {379--392},
}

@inproceedings{sener_multi-task_2018,
	address = {Montreal, QC, Canada},
	title = {Multi-task learning as multi-objective optimization},
	volume = {31},
	booktitle = {Adv {Neural} {Inf} {Process} {Syst}},
	author = {Sener, Ozan and Koltun, Vladlen},
    publisher = {Curran Associates, Inc.},
    editor = {Bengio, Samy and Wallach, Hanna and Larochelle, Hugo and Grauman, Kristen and Cesa-Bianchi, Nicol{\`o} and Garnett, Roman},
	month = dec,
	year = {2018},
	pages = {525--536},
    url = {https://papers.nips.cc/paper/2018/hash/432aca3a1e345e339f35a30c8f65edce-Abstract.html}
}

@inproceedings{venna_neighborhood_2001,
	address = {Vienna, Austria},
	title = {Neighborhood preservation in nonlinear projection methods: {An} experimental study},
	volume = {2130},
	isbn = {978-3-540-44668-2},
	doi = {10.1007/3-540-44668-0_68},
	booktitle = {Proc {ICANN}},
	publisher = {Springer},
	author = {Venna, Jarkko and Kaski, Samuel},
	editor = {Dorffner, Georg and Bischof, Horst and Hornik, Kurt},
	year = {2001},
	pages = {485--491},
}

@article{overloper_unsupervised_2026,
	title = {Unsupervised state-learning for multivariate time series data: {A} survey and practical guidelines},
	volume = {59},
	issn = {1573-7462},
	shorttitle = {Unsupervised state-learning for multivariate time series data},
	doi = {10.1007/s10462-026-11704-5},
	number = {11},
	urldate = {2026-10-23},
	journal = {Artif Intell Rev},
	author = {Overl{\"o}per, Phillip Johann and Coelho, Cec{\'i}lia and Niggemann, Oliver},
	month = sep,
	year = {2026},
	pages = {224},
}

@inproceedings{nelson_statistical_2026,
	address = {Rio de Janeiro, Brazil},
	title = {Statistical and structural identifiability in representation learning},
	author = {Nelson, Walter and Fumero, Marco and Karaletsos, Theofanis and Locatello, Francesco},
	editor = {Vondrick, Carl and Hariharan, Bharath and Raffel, Colin and Pinto, Lerrel and Yang, Diyi and Faust, Aleksandra},
    booktitle = {Proc ICLR},
	month = apr,
	year = {2026},
	pages = {151985--152020},
    url = {https://proceedings.iclr.cc/paper_files/paper/2026/hash/f67e5f99b23b108a3a6665f410034bcd-Abstract-Conference.html},
}

@inproceedings{joshi_who_2026,
	address = {Amsterdam, The Netherlands},
	title = {Who guards the guardians? {The} challenges of evaluating identifiability of learned representations},
	volume = {337},
	booktitle = {Proc {UAI}},
	publisher = {PMLR},
	author = {Joshi, Shruti and Saulus, Th{\'e}o and Brendel, Wieland and Brouillard, Philippe and Sridhar, Dhanya and Reizinger, Patrik},
	editor = {Perkovi{\'c}, Emilija and Malinsky, Daniel},
	month = aug,
	year = {2026},
	pages = {2618--2660},
    url = {https://proceedings.mlr.press/v337/joshi26a.html},
}

@inproceedings{hromadka_maximum-likelihood_2026,
	address = {Seoul, South Korea},
	title = {Maximum-likelihood learning of latent dynamics without reconstruction},
	volume = {306},
	booktitle = {Proc {ICML}},
	publisher = {PMLR},
	author = {Hromadka, Samo and Biegun, Kai and Fox, Lior and Heald, James and Sahani, Maneesh},
	editor = {Zhang, Tong and Dudik, Miroslav and Jaggi, Martin and Agarwal, Alekh and Li, Sharon and Schuurmans, Dale and Zhu, Jerry and Berkenkamp, Felix and Dong, Hanze and Bietti, Alberto},
	month = jul,
	year = {2026},
	pages = {44368--44396},
    url = {https://proceedings.mlr.press/v306/hromadka26a.html}
}

@inproceedings{simon_unsupervised_2026,
	address = {Seoul, South Korea},
	title = {Unsupervised disentanglement without compromises : {How} functional orthogonality enforces identifiability},
	volume = {306},
	booktitle = {Proc {ICML}},
	publisher = {PMLR},
	author = {Simon, Mathieu Cyrille and Frossard, Pascal and De Vleeschouwer, Christophe},
	editor = {Zhang, Tong and Dudik, Miroslav and Jaggi, Martin and Agarwal, Alekh and Li, Sharon and Schuurmans, Dale and Zhu, Jerry and Berkenkamp, Felix and Dong, Hanze and Bietti, Alberto},
	month = jul,
	year = {2026},
	pages = {113322--113359},
    url = {https://proceedings.mlr.press/v306/simon26b.html},
}

@inproceedings{zheng_diffusion_2026,
	address = {Rio de Janeiro, Brazil},
	title = {Diffusion transformers with representation autoencoders},
	booktitle = {Proc {ICLR}},
	author = {Zheng, Boyang and Ma, Nanye and Tong, Shengbang and Xie, Saining},
	editor = {Vondrick, Carl and Hariharan, Bharath and Raffel, Colin and Pinto, Lerrel and Yang, Diyi and Faust, Aleksandra},
	month = apr,
	year = {2026},
	pages = {35791--35820},
    url = {https://proceedings.iclr.cc/paper_files/paper/2026/hash/3c4141c12660ad3625eb4ae845e0a6f9-Abstract-Conference.html}
}

@inproceedings{huang_vjepa_2026,
	address = {Seoul, South Korea},
	title = {{VJEPA}: {Variational} joint embedding predictive architectures as probabilistic world models},
	volume = {306},
	booktitle = {Proc {ICML}},
	publisher = {PMLR},
	author = {Huang, Yongchao},
	editor = {Zhang, Tong and Dudik, Miroslav and Jaggi, Martin and Agarwal, Alekh and Li, Sharon and Schuurmans, Dale and Zhu, Jerry and Berkenkamp, Felix and Dong, Hanze and Bietti, Alberto},
	month = jul,
	year = {2026},
	pages = {47703--47731},
    url = {https://proceedings.mlr.press/v306/huang26ba.html},
}

@inproceedings{cheong_fairssl_2026,
	address = {Seoul, South Korea},
	title = {{FairSSL}: {Fair} multimodal self-supervised learning},
	volume = {306},
	booktitle = {Proc {ICML}},
	publisher = {PMLR},
	author = {Cheong, Jiaee and Mogharabin, Abtin and Liang, Paul Pu and Gunes, Hatice and Kalkan, Sinan},
	editor = {Zhang, Tong and Dudik, Miroslav and Jaggi, Martin and Agarwal, Alekh and Li, Sharon and Schuurmans, Dale and Zhu, Jerry and Berkenkamp, Felix and Dong, Hanze and Bietti, Alberto},
	month = jul,
	year = {2026},
	pages = {19236--19266},
    url = {https://proceedings.mlr.press/v306/cheong26a.html},
}

@inproceedings{chen_learning_2026,
	address = {Rio de Janeiro, Brazil},
	title = {Learning patient-specific disease dynamics with latent flow matching for longitudinal imaging generation},
	urldate = {2026-10-23},
	booktitle = {Proc {ICLR}},
	author = {Chen, Hao and Yin, Rui and Chen, Yifan and Chen, Qi and Li, Chao},
	month = apr,
    editor = {Vondrick, Carl and Hariharan, Bharath and Raffel, Colin and Pinto, Lerrel and Yang, Diyi and Faust, Aleksandra},
	year = {2026},
	pages = {82707--82723},
    url = {https://proceedings.iclr.cc/paper_files/paper/2026/hash/859c3ab5fe31e8d1c1a70a34fe03dde9-Abstract-Conference.html},
}

@article{hua_splitjepa_2026,
    author = {Hua, Ruijin and Liu, Zichuan and Zhao, Zhuokai and Zheng, Yujia},
    title = {{SplitJEPA}: {Learning} invariant and variant latent worlds without reconstruction},
    journal = {Preprint at arXiv},
    year = {2026},
    doi = {10.48550/arXiv.2610.12349}
}


\newpage

\section*{FIGURE LEGENDS}

\refstepcounter{figure}
\label{fig:tradeoffs}
\noindent Figure~\thefigure. \textbf{Interactions among latent-state constraints.}
Predictive sufficiency, minimality, and nuisance invariance define a qualitative projection of a broader design space. Model families are positioned according to the properties they typically emphasize. Links indicate possible interactions, whose nature depends on the task and modeling assumptions, rather than universal conflicts.

\bigskip

\refstepcounter{figure}
\label{fig:taxonomy}
\noindent Figure~\thefigure. \textbf{Taxonomy of latent representation models.}
Six broad families of approaches and their representative subcategories are shown. The taxonomy is schematic, neither exhaustive nor mutually exclusive, and combines methodological and domain-specific perspectives. It highlights recurring design choices rather than fixed properties shared by every model within a family.

\bigskip

\refstepcounter{figure}
\label{fig:pipeline}
\noindent Figure~\thefigure. \textbf{Representation design under complementary constraints.}
Six complementary properties guide the learning of latent states from observations. Their relative importance and implementation depend on the task and modeling assumptions. Learned representations can then support prediction, interpretation, and simulation.

\bigskip

\refstepcounter{figure}
\label{fig:pareto_fronts}
\noindent Figure~\thefigure. \textbf{Empirical pairwise Pareto fronts obtained by varying the six training-objective weights.}
(A) Raw-observation prediction MSE ($h=5$, $\downarrow$) versus counterfactual relative energy ($\downarrow$).
(B) Raw-observation prediction MSE ($h=5$, $\downarrow$) versus nuisance $Z$ balanced accuracy ($\downarrow$).
(C) Linear state $R^2$ ($\uparrow$) versus counterfactual relative energy ($\downarrow$).
(D) Linear state $R^2$ ($\uparrow$) versus neighborhood trustworthiness ($\uparrow$).
The empirical fronts are based on 200 evaluated weight configurations, each trained with five independent random seeds. Each point represents the mean validation performance of one configuration across seeds; orange points identify non-dominated configurations for the corresponding pair of metrics, and staircase curves connect these Pareto-optimal solutions. Labels identify Pareto-optimal configurations, whose constraint weights and panel memberships are reported for a representative subset in Table~\ref{tab:pareto-weights}. See also Table~S1.

\bigskip

\refstepcounter{figure}
\label{fig:environment_representation}
\noindent Figure~\thefigure. \textbf{Latent geometry and temporal organization.}
Latent representations in panels B--D are obtained from a single model trained with configuration P29.
(A) Two nuisance-conditioned observations of the same physical trajectory in measurement space.
(B) PCA projection of the learned latent representations, colored by the true horizontal position $x$.
(C) Counterfactual observations projected into the same latent space.
(D) One latent trajectory shown in the same PCA coordinates, colored by time.
Panels B--D share the same projection and axis limits.

\bigskip

\refstepcounter{figure}
\label{fig:identifiability}
\noindent Figure~\thefigure. \textbf{Coordinate ambiguity and constraint-dependent regularization.}
Orthogonal reparameterizations of the four-dimensional latent space of a single model trained with configuration P29.
(A,B) Original and rotated representations, projected onto the fixed reference PCA axes and colored by the true horizontal position $x$.
(C) Numerical verification of the algebraic invariance of reconstruction, five-step prediction, and temporal-coherence errors under consistent orthogonal reparameterization.
(D) Mean absolute latent activation and VICReg-inspired structural penalties vary with rotation angle, shown relative to their values at $0^\circ$.
Coordinate-dependent regularization can favor particular parameterizations without establishing identifiability of the underlying physical factors. See also Figure~S2.


\newpage

\section*{FIGURES}

\begin{center}
  Figure 1\\[0.5cm]
  \includegraphics[width=0.7\textwidth]{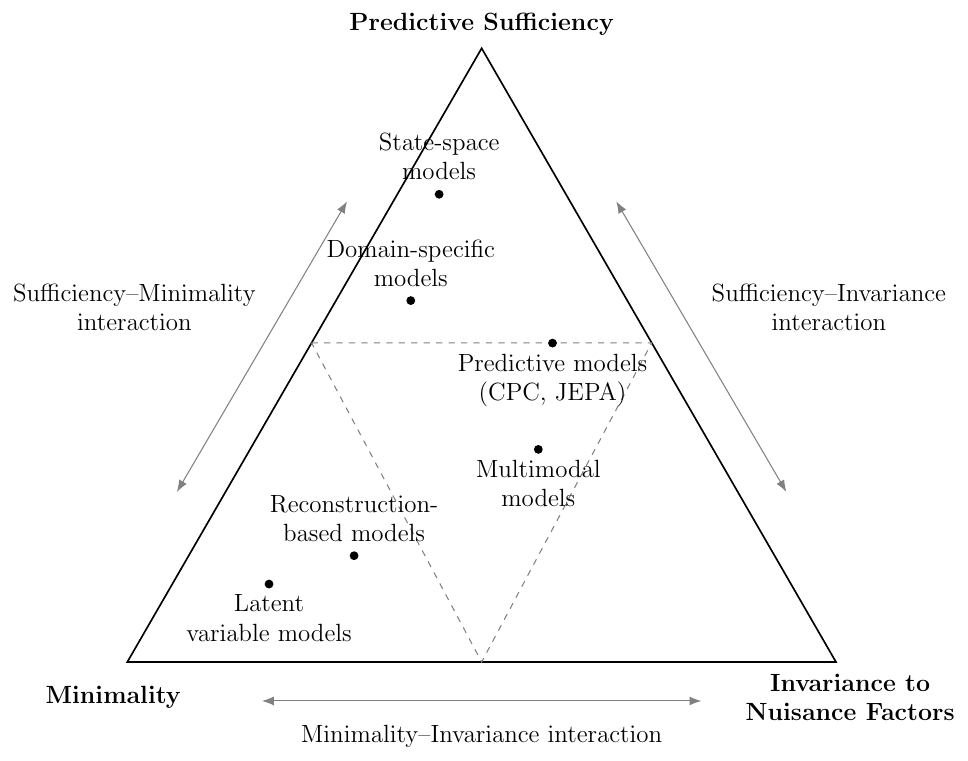}
\end{center}

\newpage

\begin{center}
  Figure 2\\[0.5cm]
  \includegraphics[width=0.8\textwidth]{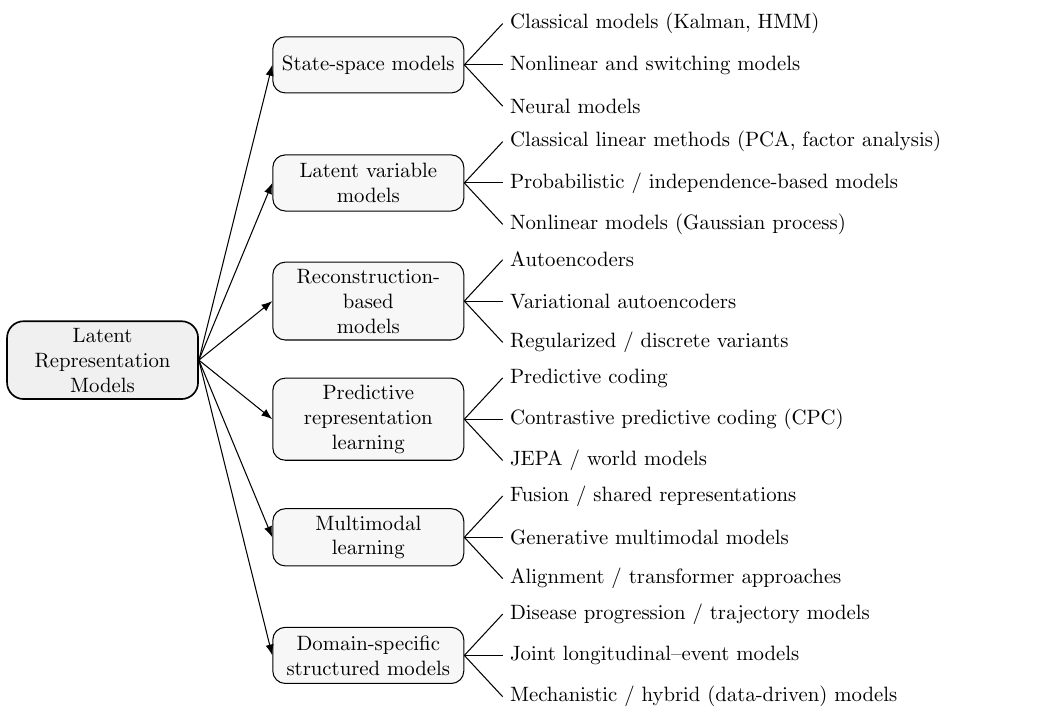}
\end{center}

\newpage

\begin{center}
  Figure 3\\[0.5cm]
  \includegraphics[width=0.8\textwidth]{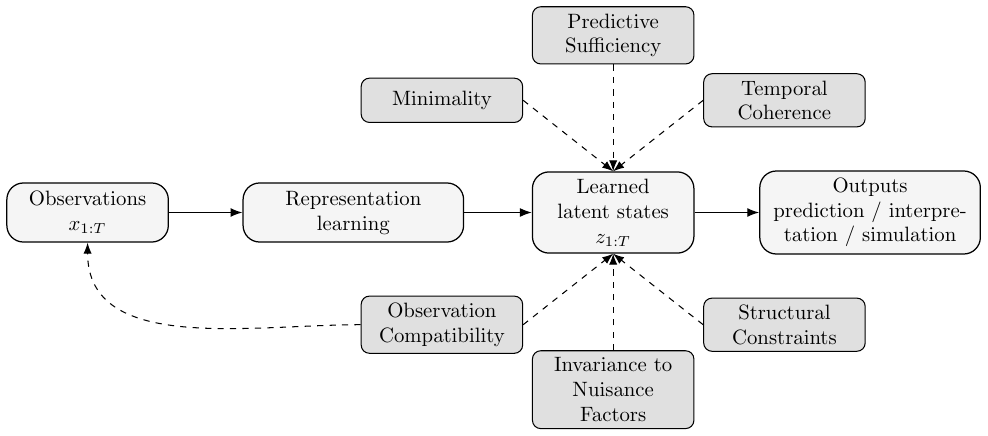}
\end{center}

\newpage

\begin{center}
  Figure 4\\[0.5cm]
  \includegraphics[width=\textwidth]{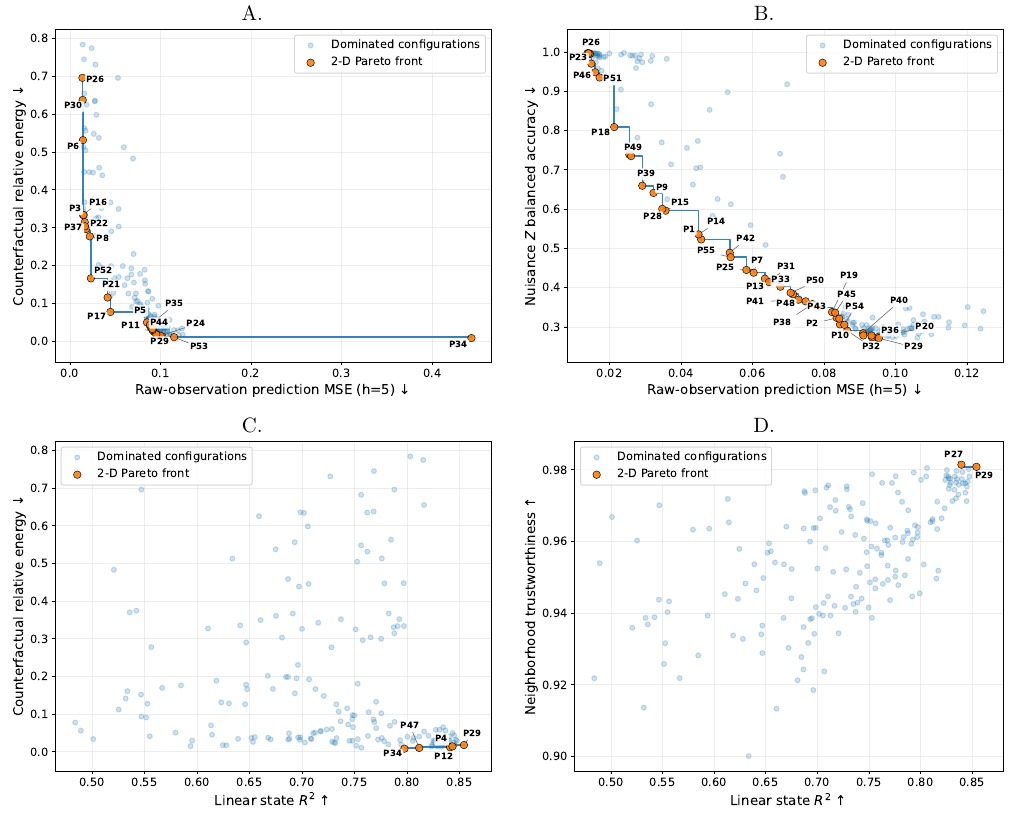}
\end{center}

\newpage

\begin{center}
  Figure 5\\[0.5cm]
  \includegraphics[width=\textwidth]{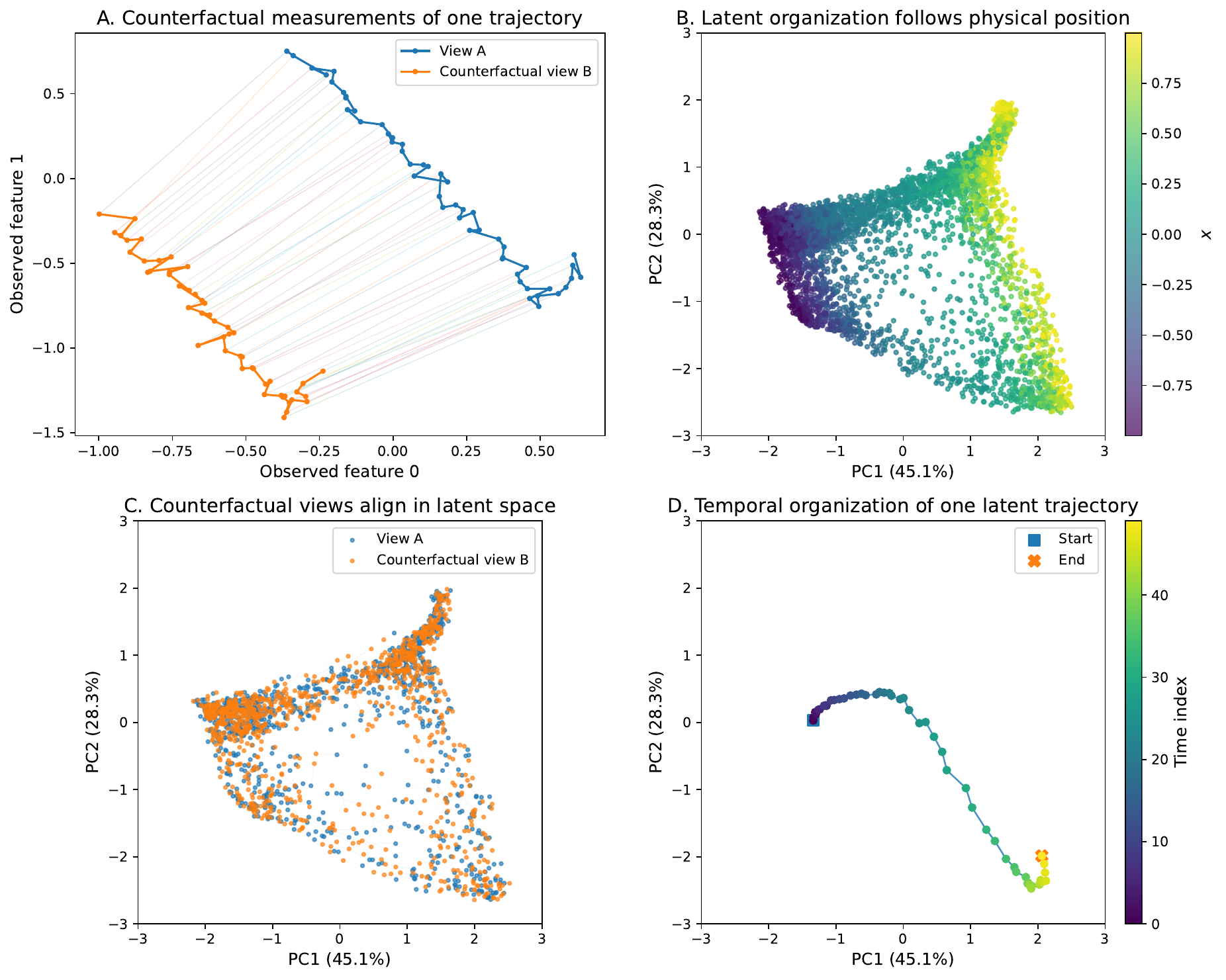}
\end{center}

\newpage

\begin{center}
  Figure 6\\[0.5cm]
  \includegraphics[width=\textwidth]{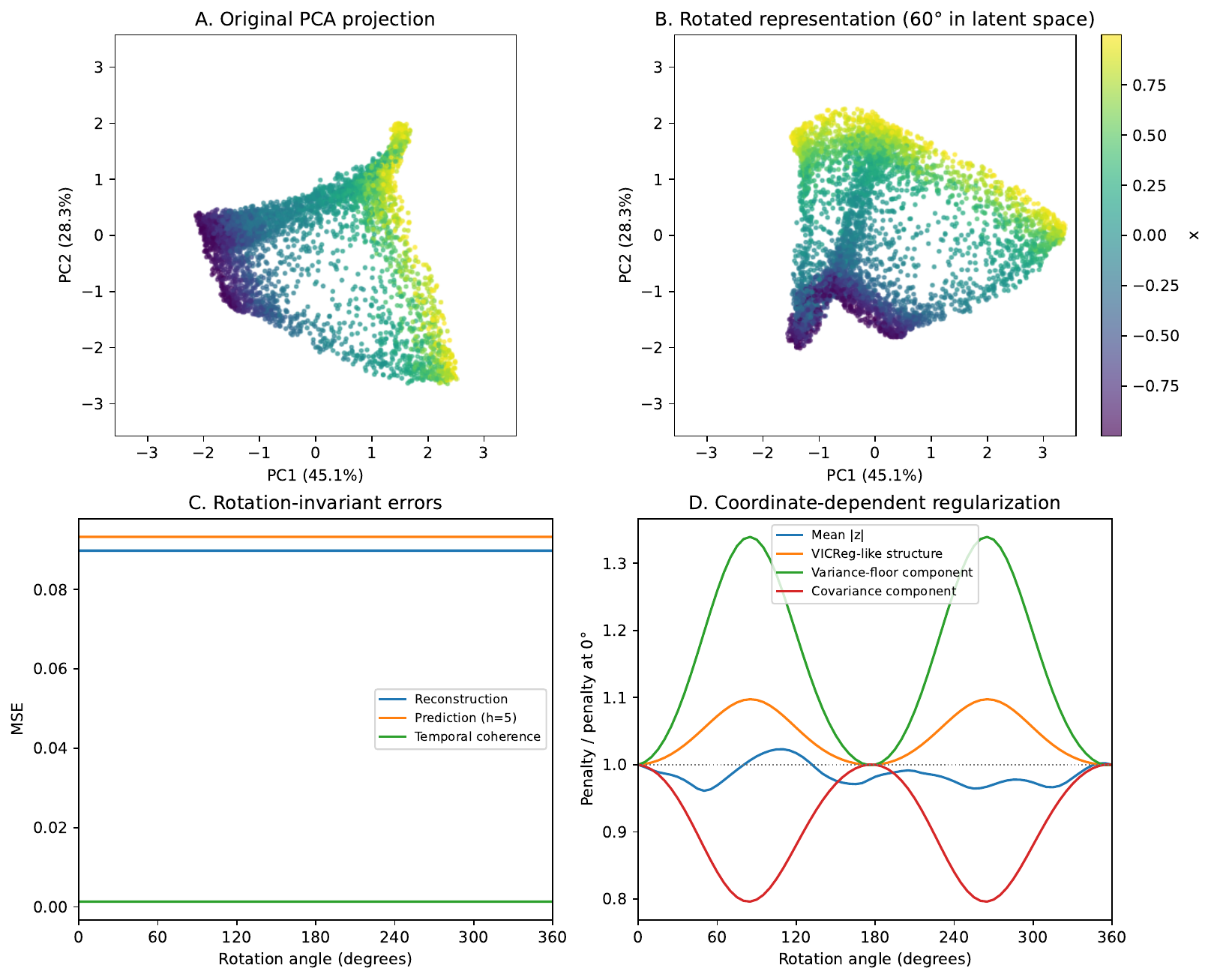}
\end{center}


\clearpage

\section*{TABLES}

\begin{table}[h]
  \centering
  \caption{
    \textbf{Representation properties emphasized by major modeling families.}
    Symbols indicate typical design priorities: $\checkmark$, explicitly targeted in canonical formulations through objectives, probabilistic assumptions, or architectural constraints; $\sim$, indirectly encouraged, partially addressed, or targeted in common variants; $\times$, not a characteristic objective. Assignments are qualitative and may differ across implementations. Pred = Predictive Sufficiency; Min = Minimality; Temp = Temporal Coherence; Obs = Observation Compatibility; Inv = Invariance to Nuisance Factors; Struct = Structural Constraints.
  }
  \label{tab:comparison}
  \small
  \begin{tabular}{lcccccc}
    \toprule
    \textbf{Model family} & \textbf{Pred} & \textbf{Min} & \textbf{Temp} & \textbf{Obs} & \textbf{Inv} & \textbf{Struct} \\
    \midrule
    State-space models & $\checkmark$ & $\sim$ & $\checkmark$ & $\checkmark$ & $\times$ & $\times$ \\
    Latent variable models & $\times$ & $\checkmark$ & $\times$ & $\checkmark$ & $\times$ & $\checkmark$ \\
    Reconstruction-based models & $\times$ & $\checkmark$ & $\times$ & $\checkmark$ & $\times$ & $\times$ \\
    Predictive representation learning & $\checkmark$ & $\sim$ & $\sim$ & $\times$ & $\sim$ & $\times$ \\
    Multimodal learning & $\sim$ & $\times$ & $\times$ & $\sim$ & $\sim$ & $\times$ \\
    Domain-specific models & $\sim$ & $\times$ & $\checkmark$ & $\checkmark$ & $\times$ & $\checkmark$ \\
    \bottomrule
  \end{tabular}
\end{table}


\bigskip

\begin{table}[h]
  \centering
  \caption{\textbf{Characteristics of the synthetic benchmark.}}
  \label{tab:toy_dataset}
  \small
  \begin{tabular}{lc}
    \toprule
    \textbf{Property}         & \textbf{Value} \\
    \midrule
    Latent state dimension    & 4 \\
    Observation dimension     & 12 channels \\
    Sequence length           & 50 time steps \\
    Training trajectories     & 5,000 \\
    Validation trajectories   & 1,000 \\
    Test trajectories         & 1,000 \\
    OOD trajectories          & 1,000 \\
    ID nuisance conditions    & 4 \\
    ID nuisance-class prior   & Uniform ($25\%$ each) \\
    OOD nuisance distribution & Shifted two-component mixture \\
    \bottomrule
  \end{tabular}
\end{table}


\clearpage

\begin{table}[h]
  \centering
  \caption{\textbf{Experimental and sweep hyperparameters.}}
  \label{tab:hyperparameters}
  \small
  \setstretch{1.05}
  \renewcommand{\arraystretch}{1.2}
  \begin{tabular}{
    @{}
    >{\raggedright\arraybackslash}p{0.09\linewidth}
    >{\raggedright\arraybackslash}p{0.17\linewidth}
    >{\raggedright\arraybackslash}p{0.17\linewidth}
    >{\raggedright\arraybackslash}p{0.49\linewidth}
    @{}
  }
    \toprule
    \textbf{Category} & \textbf{Parameter} & \textbf{Value / range} & \textbf{Notes} \\
    \midrule

    Sweep
      & Configurations
      & 200
      & 20 reference, 120 targeted, 60 exploratory \\[2pt]

      & Sweep seed
      & 12,345
      & Fixed \\

    \midrule

    Weights
      & $\lambda_{\mathrm{pred}}$, $\lambda_{\mathrm{temp}}$
      & $[0.01,5]$, $[0.01,2]$
      & Log-uniform when positive \\

      & $\lambda_{\mathrm{obs}}$, $\lambda_{\mathrm{inv}}$
      & $[0.003,3]$, $[0.01,5]$
      & Log-uniform when positive \\

      & $\lambda_{\mathrm{min}}$, $\lambda_{\mathrm{struct}}$
      & $[10^{-5},5\times10^{-2}]$, $[10^{-3},1]$
      & Log-uniform when positive \\

      & Sampling design
      & Fixed zero quotas; $\pm0.5$ log$_{10}$
      & One or two zero weights; targeted samples truncated to global ranges \\

    \midrule

    Training
      & Model seeds
      & 0, 1, 2, 3, 4
      & Five runs per configuration \\

      & Trainable parameters
      & 16,020 backbone; +4,740 adversary
      & 20,760 total when adversarial invariance is active \\

      & Optimizers
      & AdamW / Adam
      & Backbone: lr $10^{-3}$, wd $10^{-5}$; adversary: lr $10^{-3}$, wd $10^{-4}$ \\

      & Training budget
      & 50 epochs; batch 128
      & Fixed final checkpoint; no early stopping; gradient clipping 5 \\

      & Adversarial warmup
      & 20 epochs
      & Linear increase to the configured invariance strength \\

      & Refresh phase
      & Epochs 41--50
      & Adversary is trained on a refreshed pool of 25,000 training latent codes instead of the current minibatch; normalization statistics are fixed \\

      & Adversary updates
      & 15 per backbone step
      & 512 latent positions per update \\

    \midrule

    Evaluation
      & Evaluation batch size
      & 256
      & \\

      & Rollout horizons
      & 1, 5, 10
      & $h=5$ for Pareto analysis \\

      & Nuisance probe
      & 500 epochs; seed 42
      & Fixed probe seed \\

      & Nonlinear probe training samples
      & Up to 25,000 time points
      & Drawn from training trajectories only \\

    \bottomrule
  \end{tabular}
\end{table}


\clearpage

\begin{table}[h]
  \centering
  \small
  \caption{\textbf{Constraint weights of representative configurations on the empirical pairwise Pareto fronts.}
  Panel membership refers to the validation fronts.}
  \label{tab:pareto-weights}
  \begin{tabular}{lrrrrrrl}
    \hline
    Configuration & $\lambda_{\rm pred}$ & $\lambda_{\rm min}$ & $\lambda_{\rm temp}$ & $\lambda_{\rm obs}$ & $\lambda_{\rm inv}$ & $\lambda_{\rm struct}$ & Panels \\
    \hline
    P1 & 1 & 0 & 0.5 & 0.5 & 0.1 & 0 & (B) \\
    P2 & 1 & 0 & 0.5 & 0.5 & 0.3 & 0 & (B) \\
    P5 & 0.0610 & $5.38\times10^{-5}$ & 0.856 & 0 & 0.0100 & 0.0112 & (A) \\
    P7 & 0.314 & 0 & 1.45 & 0.00584 & 0.0295 & 0.0154 & (B) \\
    P12 & 0.0247 & $3.62\times10^{-5}$ & 0.350 & 0 & 0.0331 & 0.00717 & (C) \\
    P17 & 0.0747 & $7.64\times10^{-5}$ & 0.927 & 2.11 & 0.0677 & 0.00724 & (A) \\
    P18 & 0.582 & 0 & 0.613 & 0 & 0.0117 & 0.0159 & (B) \\
    P22 & 1.50 & $1.90\times10^{-5}$ & 0.0156 & 1.31 & 0.0253 & 0 & (A) \\
    P26 & 2.85 & 0.000586 & 1.600 & 0.192 & 0 & 0.0554 & (A), (B) \\
    P27 & 0.0791 & $1.53\times10^{-5}$ & 0.271 & 0.0310 & 0.0649 & 0.00792 & (D) \\
    P29 & 0.0182 & $4.62\times10^{-5}$ & 0.352 & 0 & 0.0109 & 0.00518 & (A), (B), \\
        &        &          &       &   &        &         & (C), (D) \\
    P34 & 0 & $3.73\times10^{-5}$ & 1.29 & 0 & 0.027 & 0.013 & (A), (C) \\
    \hline
  \end{tabular}
\end{table}


\clearpage

\begin{table*}[h]
  \centering
  \small
  \renewcommand{\arraystretch}{0.88}
  \setlength{\tabcolsep}{4pt}
  \caption
  {
    \textbf{Performance of selected training configurations.}
    Values are means across model seeds. Arrows indicate the preferred numerical direction where applicable. Low nuisance decodability or counterfactual energy may reflect collapsed representations, mean $|z|$ measures activation magnitude rather than information-theoretic minimality, and unnormalized latent-transition MSE is not comparable across configurations with different latent scales. CF: counterfactual; BA: balanced accuracy; OOD: out-of-distribution.
  }
  \label{tab:main-results}
  \begin{tabular}{lcccccc}
    \toprule
    Configuration & \makecell{Raw-obs. Pred. \\ MSE $h=5 \downarrow$} & \makecell{Linear \\ state $R^2 \uparrow$} & \makecell{Neigh. \\ Trust. $\uparrow$} & \makecell{CF rel. \\ energy $\downarrow$} & \makecell{Nuis. $Z$ \\ BA $\downarrow$} & \makecell{Nuis. $(S,Z)$ \\ BA $\downarrow$} \\
    \midrule
    predictive & 0.0168 & 0.631 & 0.945 & 0.341 & 0.988 & 0.995 \\
    reconstruction & 0.643 & 0.692 & 0.949 & 0.334 & 0.990 & 0.996 \\
    temporal & 0.427 & 0.0916 & 0.623 & 1.80 & 0.511 & 0.807 \\
    structural & 0.473 & 0.412 & 0.862 & 0.688 & 0.827 & 0.959 \\
    P1 & 0.0449 & 0.646 & 0.929 & 0.291 & 0.524 & 0.855 \\
    P2 & 0.0823 & 0.751 & 0.958 & 0.0879 & 0.314 & 0.760 \\
    P5 & 0.0834 & 0.835 & 0.972 & 0.0515 & 0.320 & 0.765 \\
    P7 & 0.0593 & 0.718 & 0.934 & 0.146 & 0.442 & 0.845 \\
    P12 & 0.0965 & 0.834 & 0.976 & 0.0149 & 0.296 & 0.453 \\
    P17 & 0.0451 & 0.486 & 0.919 & 0.0796 & 0.752 & 0.880 \\
    P18 & 0.0219 & 0.769 & 0.945 & 0.456 & 0.791 & 0.930 \\
    P22 & 0.0179 & 0.748 & 0.955 & 0.304 & 0.945 & 0.982 \\
    P26 & 0.0139\textsuperscript{a} & 0.779 & 0.954 & 0.715 & 0.995 & 0.996 \\
    P27 & 0.0909 & 0.833 & 0.981\textsuperscript{a} & 0.0303 & 0.288 & 0.570 \\
    P29 & 0.0931 & 0.849\textsuperscript{a} & 0.980 & 0.0183 & 0.272 & 0.526 \\
    P34 & 0.437 & 0.790 & 0.963 & 0.00860 & 0.284 & 0.346 \\
    \bottomrule
  \end{tabular}

  \vspace{0.75em}

  \begin{tabular}{lccccc}
    \toprule
    Configuration & \makecell{Obs. recon. \\ MSE $\downarrow$} & \makecell{1-step latent \\ trans. MSE} & \makecell{Mean $|z|$} & \makecell{OOD raw-obs. \\ pred. MSE $h=5 \downarrow$} & \makecell{OOD linear \\ state $R^2 \uparrow$} \\
    \midrule
    predictive & 0.0363 & 0.255 & 0.775 & 0.0836 & 0.421 \\
    reconstruction & 0.0126\textsuperscript{a} & 0.231 & 0.460 & 0.970 & 0.570 \\
    temporal & 0.426 & $7.84\times10^{-6}$ & 0.517 & 0.947 & -0.175 \\
    structural & 0.467 & 0.210 & 0.891 & 0.994 & 0.256 \\
    P1 & 0.0428 & 0.00289 & 0.608 & 0.247 & 0.492 \\
    P2 & 0.0799 & 0.00245 & 0.562 & 0.500 & 0.647 \\
    P5 & 0.0808 & 0.00140 & 0.780 & 0.499 & 0.767 \\
    P7 & 0.0585 & 0.00155 & 0.725 & 0.323 & 0.543 \\
    P12 & 0.0942 & 0.00208 & 0.858 & 0.607 & 0.820 \\
    P17 & 0.0249 & 0.00222 & 0.584 & 0.160 & 0.456 \\
    P18 & 0.0244 & 0.00341 & 0.925 & 0.119 & 0.673 \\
    P22 & 0.0148 & 0.00397 & 0.669 & 0.0830 & 0.638 \\
    P26 & 0.0169 & 0.00342 & 0.996 & 0.0758\textsuperscript{a} & 0.712 \\
    P27 & 0.0881 & 0.00266 & 0.779 & 0.581 & 0.808 \\
    P29 & 0.0894 & 0.00130 & 0.771 & 0.583 & 0.830\textsuperscript{a} \\
    P34 & 0.437 & 0.000841 & 0.772 & 0.956 & 0.778 \\
    \bottomrule
  \end{tabular}

  \vspace{0.5em}

  \begin{minipage}{\textwidth}
    \footnotesize
    \textsuperscript{a} Best numerical value among the reported configurations for the corresponding diagnostic, according to the indicated preferred direction. Best values are not highlighted for diagnostics whose interpretation requires additional caution.
  \end{minipage}
\end{table*}


\clearpage

\section*{SUPPLEMENTARY INFORMATION}

\noindent Table S1. \textbf{Leave-one-seed-out stability of empirical pairwise Pareto fronts.}

\begin{center}
  \small
  \begin{tabular}{lrrr}
    \hline
    Front & $N_{\mathrm{front}}$ & Jaccard: mean [min--max] & Retained \% ($n/N$) \\
    \hline
    (a) & 28 & $0.697$ [0.636--0.765] & 80.7\% (113/140) \\
    (b) & 42 & $0.779$ [0.766--0.813] & 87.1\% (183/210) \\
    (c) & 5 & $0.706$ [0.429--0.833] & 80.0\% (20/25) \\
    (d) & 2 & $0.700$ [0.333--1.000] & 90.0\% (9/10) \\
    \hline
  \end{tabular}
\end{center}

\noindent\small
Fronts (a–d) correspond to the validation-data metric pairs shown in Figure~4 of the main manuscript. Each of the five analyses excludes one model seed. Jaccard values are the mean and observed range (minimum--maximum) of similarity to the original five-seed front. Retention is the percentage and total number of original-front memberships preserved across the five analyses, divided by $5\times N_{\mathrm{front}}$. Memberships are counted repeatedly across omissions and do not represent independent experiments.
\normalsize

\bigskip

\noindent Table S2. \textbf{Effect of prediction targets and adversarial invariance.}

\begin{center}
  \small
  \begin{tabular}{@{}lccc@{}}
    \toprule
    Metric & Original P29 & P29--6 & P29--6 without \\
           &              &        & adversarial invariance \\
    \midrule
    \multicolumn{4}{@{}l}{\emph{Representation and invariance}} \\
    Nuisance BA from $Z$ (\%) & $27.22 \pm 0.65$ & $25.69 \pm 0.40$ & $53.11 \pm 37.09$ \\
    Nuisance BA from $(S,Z)$ (\%) & $52.60 \pm 2.46$ & $25.88 \pm 0.45$ & $56.15 \pm 39.71$ \\
    BA gain over $S$ (percentage points) & $26.10 \pm 2.46$ & $-0.63 \pm 0.45$ & $29.64 \pm 39.71$ \\
    Continuous nuisance $R^2$ from $(S,Z)$ & $0.3753 \pm 0.0349$ & $0.0165 \pm 0.0057$ & $0.3445 \pm 0.4244$ \\
    Linear physical-state $R^2$ & $0.849 \pm 0.034$ & $0.685 \pm 0.160$ & $0.634 \pm 0.220$ \\
    Nonlinear future-state $R^2$ ($h=5$) & $0.954 \pm 0.005$ & $0.941 \pm 0.020$ & $0.879 \pm 0.082$ \\
    Counterfactual relative energy (ID) & $0.0183 \pm 0.0019$ & $0.0072 \pm 0.0012$ & $0.4236 \pm 0.5725$ \\
    \addlinespace[3pt]
    \multicolumn{4}{@{}l}{\emph{Prediction on six common channels}} \\
    Six-channel rollout MSE ($h=5$, ID) & $0.02016 \pm 0.00425$ & $0.01381 \pm 0.00198$ & $0.01748 \pm 0.00511$ \\
    Six-channel rollout MSE ($h=5$, OOD) & $0.03180 \pm 0.00350$ & $0.02240 \pm 0.00305$ & $0.03376 \pm 0.01505$ \\
    \bottomrule
  \end{tabular}
\end{center}

\noindent\small
Values are mean $\pm$ sample standard deviation over five training seeds (0--4), evaluated on the ID test set unless OOD is specified. BA denotes balanced accuracy (chance: 25\%); BA gain is the signed difference between the joint $(S,Z)$ and $S$-only strong nuisance probes, in percentage points. $S$ is the simulated physical state and $Z$ the learned latent representation. All variants receive the same 12-channel input. Original P29 predicts all 12 channels, whereas P29--6 predicts only six nuisance-independent channels (zero-based indices 2--7), retaining the original loss weights. The no-invariance variant sets $\lambda_{\mathrm{inv}}=0$, leaving other weights unchanged. Prediction MSE is evaluated on the same six channels for all variants. Target restriction also changes the effective predictive-loss normalization. Without adversarial invariance, two of five runs exceeded 99\% joint nuisance BA, revealing strong seed dependence. Physical states are used only for post-hoc evaluation; nuisance labels are also used during adversarial training in the invariant configurations.
\normalsize

\clearpage

\noindent Figure S1. \textbf{Sensitivity of nuisance probes to synthetic nuisance leakage.}

\begin{center}
  \includegraphics[width=\textwidth]{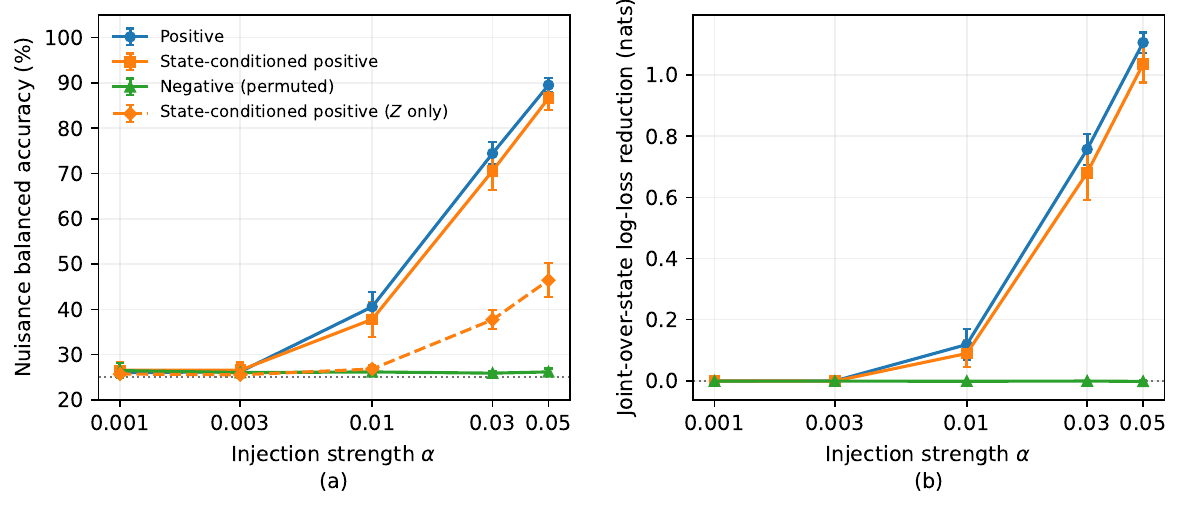}
\end{center}
\vspace{-0.4cm}

\noindent\small
Sensitivity of strong post-hoc nuisance probes to synthetic leakage injected into P29--6 latent representations at increasing strengths $\alpha$, scaled by training-set latent standard deviations. Positive controls inject either a fixed nuisance-dependent code or a state-dependent modulation of that code; the negative control uses nuisance labels independently permuted within each split. Probes are fitted on training data, selected on validation data, and evaluated on the held-out ID test set.
\textbf{(a)} Nuisance balanced accuracy from $(S,Z)$ for all controls, with an additional $Z$-only probe for the state-conditioned positive control (chance: 25\%).
\textbf{(b)} Reduction in test-set nuisance classification log-loss relative to the $S$-only probe, $\mathrm{CE}(N\mid S)-\mathrm{CE}(N\mid S,Z)$, estimated using the fitted probes, where $N$ denotes the categorical nuisance label.
Points and error bars show mean $\pm$ sample standard deviation across five training seeds. At $\alpha=0.01$, joint BA reached $40.57\pm3.21\%$ and $37.79\pm3.86\%$ for the direct and state-conditioned positive controls, respectively, versus $26.17\pm0.93\%$ for the negative control. For the state-conditioned control, $Z$-only BA remained near chance ($26.82\pm0.95\%$). Log-loss reductions showed the same trend. These results demonstrate sensitivity to synthetic nuisance leakage, including signals more readily decoded jointly with the physical state, without establishing exact conditional independence.
\normalsize

\clearpage

\noindent Figure S2. \textbf{Cross-seed reproducibility of learned latent representations.}

\begin{center}
  \includegraphics[width=\textwidth]{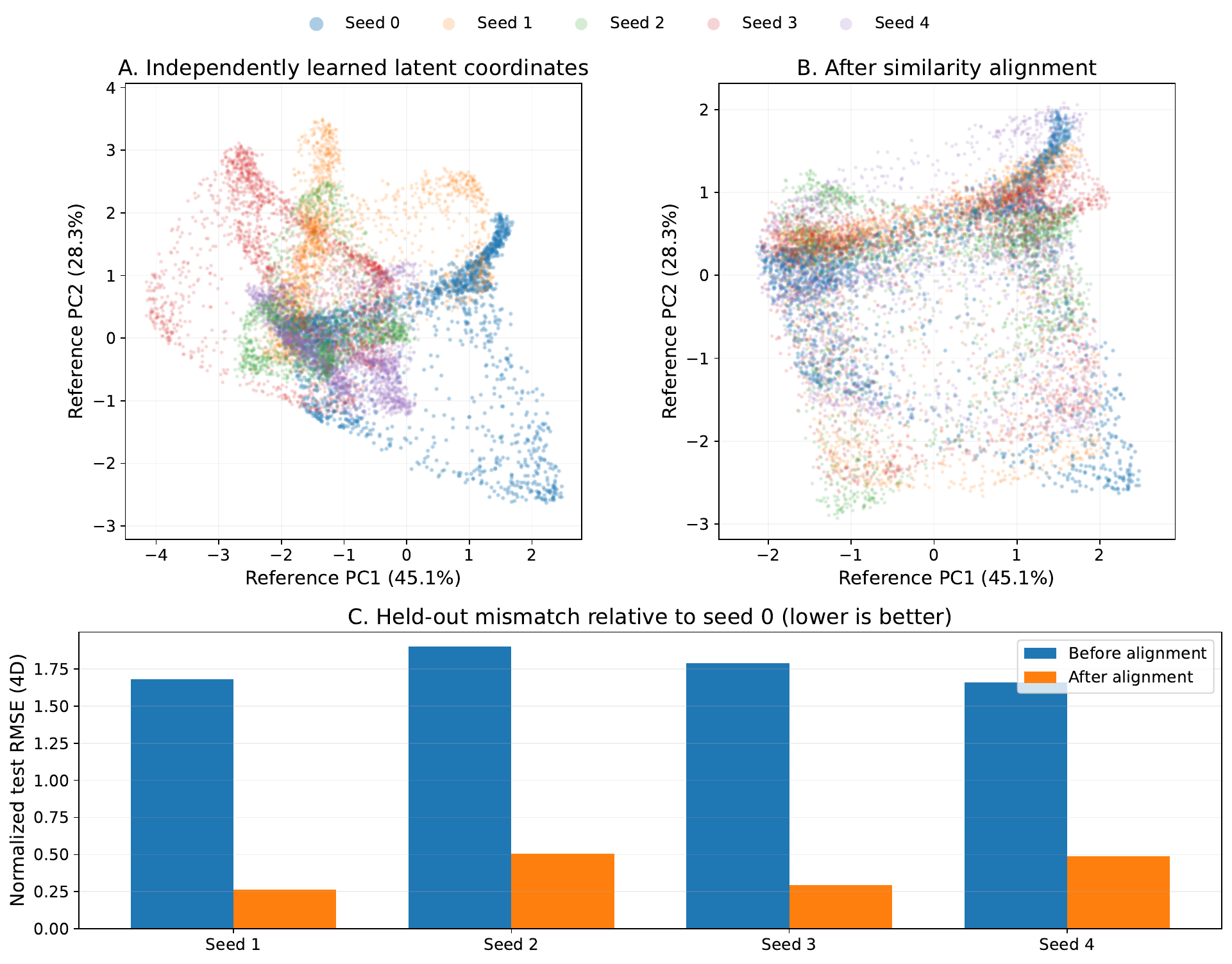}
\end{center}

\noindent\small
Comparison of four-dimensional latent representations learned independently from five training seeds of P29.
\textbf{(A)} Unaligned representations projected into the reference PCA space of seed 0.
\textbf{(B)} The same representations after similarity alignment to seed 0.
\textbf{(C)} Normalized held-out test RMSE relative to seed 0, before and after alignment.
Transformations were estimated on validation data and evaluated on held-out test data. Across all ten seed pairs, orthogonal alignment reduced normalized mismatch by 77.1\% on average, compared with 78.3\% when an additional scaling factor was allowed. Residual mismatch indicates that the representations are not exactly equivalent under these transformations.
\normalsize

\end{document}